\pdfoutput=1
\documentclass[11pt]{article}
\usepackage{ACL2023}
\input{config.sty}
\input{glossary.sty}

\title{AutoDSL: Automated domain-specific language design for structural representation of procedures with constraints}

\author{
Yu-Zhe Shi$^{\star{}}$, Haofei Hou$^{\star{}}$, Zhangqian Bi, Fanxu Meng, Xiang Wei,\\ 
\textbf{Lecheng Ruan$^{\textrm{\Letter}}$, Qining Wang$^{\textrm{\Letter}}$}
\vspace{0.5em}\\
Department of Advanced Manufacturing and Robotics, \\College of Engineering, Peking University \quad{}\\
$^\star{}$Equal contribution\quad
$\textrm{\Letter}$\phantom\,\,\texttt{\{ruanlecheng, qiningwang\}@pku.edu.cn}}

\begin{document}
\maketitle

\begin{abstract}
Accurate representation of procedures in restricted scenarios, such as non-standardized scientific experiments, requires precise depiction of constraints. Unfortunately, \ac{dsl}, as an effective tool to express constraints structurally, often requires case-by-case hand-crafting, necessitating customized, labor-intensive efforts. To overcome this challenge, we introduce the \ad framework to automate \ac{dsl}-based constraint design across various domains. Utilizing domain specified experimental protocol corpora, \ad optimizes syntactic constraints and abstracts semantic constraints. Quantitative and qualitative analyses of the \acp{dsl} designed by \ad across five distinct domains highlight its potential as an auxiliary module for language models, aiming to improve procedural planning and execution.
\end{abstract}

\section{Introduction}

Comprehending and executing procedures articulated in natural language to achieve a specified goal represents a fundamental challenge for \ac{ai} systems. With the boost of \ac{llm}~\citep{brown2020language,raffel2020exploring, touvron2023llama}, \ac{ai} systems possess the capability of reasoning over and planning for procedural tasks intended for both human and robotic execution across a broad spectrum of everyday scenarios\footnote{Visit \url{www.wikiHow.com} for demonstrations.}~\citep{pareti2014semantic,pareti2014integrating,tandon2020dataset,donatelli2021aligning,zhou2022show}, such as cooking according to a recipe, obviating the necessity for external representation of procedures beyond text.

\begin{figure}[t!]
    \centering
    \includegraphics[width=\linewidth]{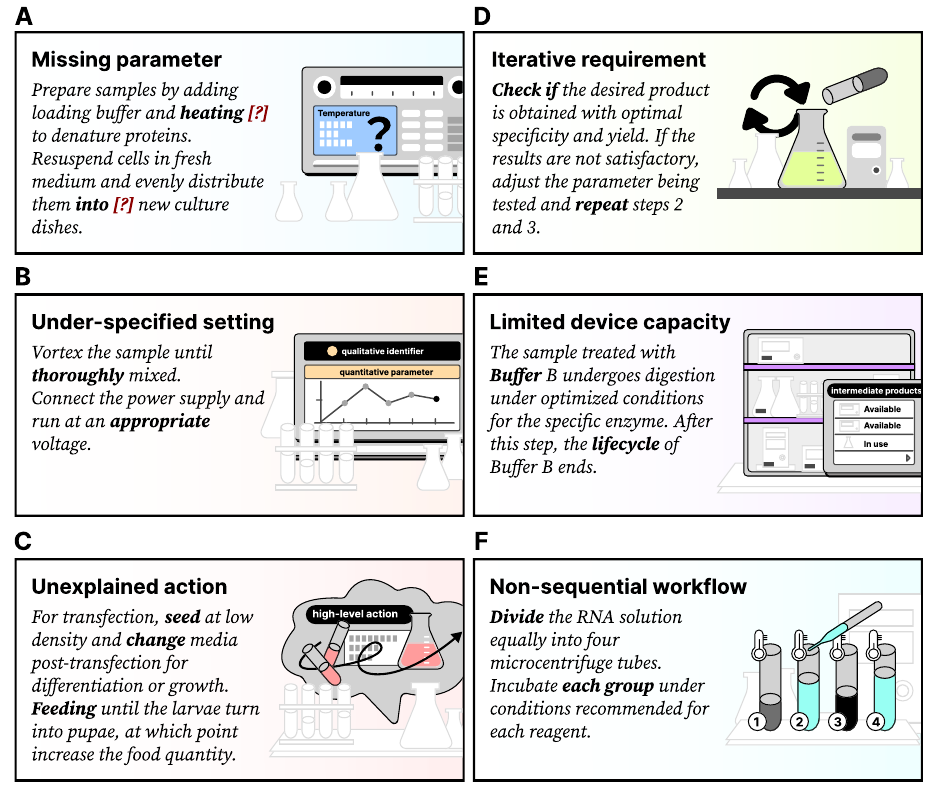}
    \caption{\textbf{Representative constraints in protocols.} 
    \textbf{(A)} Parameter omission: This refers to the absence of essential parameter values within a predefined set, \eg, the lack of temperature specification during the denaturation step in Protein Gel Electrophoresis. \textbf{(B)} Parameter under-specification: This occurs when a quantitative parameter is described using qualitative terms, leading to ambiguity, \eg, unclear mixture configurations in DNA Extraction. \textbf{(C)} Action undefinition: This involves the description of procedural steps at a high level without grounding to the specific, executable actions required, \eg, the vague \texttt{change} operation in Cell Preparation. \textbf{(D)} Iterative control logic: Loops that operate iteratively to satisfy a final condition rather than straightforwardly, as seen in PCR Optimization. \textbf{(E)} Memory management: Drawing a parallel with computer memory mechanisms, laboratory procedures also face constraints on the availability of storage for intermediates, necessitating explicit reallocation of containers and devices, such as managing buffers in Protein Digestion. \textbf{(F)} Concurrent management: The synchronization of actions without dependencies to maximize time efficiency and resource utilization, \eg, reagent splitting in RNA Extraction.}
    \label{fig:Fig1_example}
\end{figure}

\begin{figure}[t!]
    \centering
    \includegraphics[width=\linewidth]{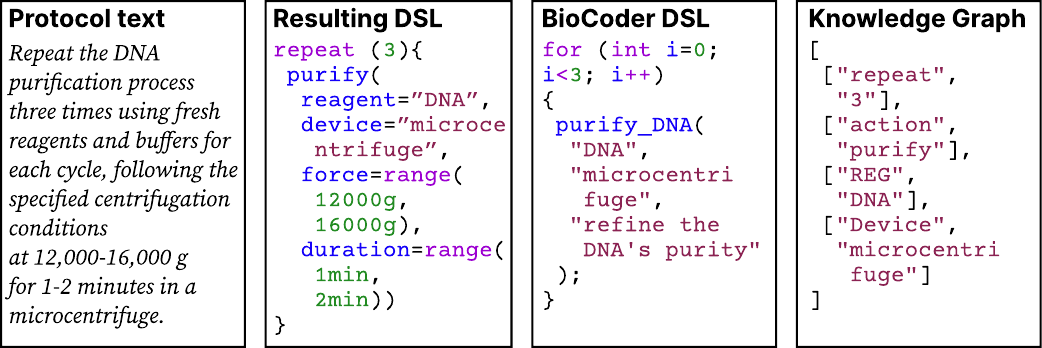}
    \caption{\textbf{Protocols in different structural representations}}
    \label{fig:Fig1.5_program_examples}
\end{figure}

However, contexts significantly more restricted than everyday scenarios, such as conducting non-standardized experiments in scientific laboratories, need to follow specific \emph{protocols}\footnote{Visit \url{www.nature.com/nprot/} for examples.}. A protocol delineates every aspect of an experiment's procedure to facilitate its reproduction~\citep{baker20161,munafo2017manifesto}, emphasizing the necessity for precision in every step, to ensure accurate execution by an experimenter. The complexity of procedures, limitation in resources, and susceptibility to error in such scenarios render any deviation from the established protocols inadmissible. Unfortunately, natural language inherently possesses ambiguities~\citep{russell1923vagueness}. Within protocols, description of actions can be semantically under-specified, and the logic of procedure may be non-linear, as demonstrated in \cref{fig:Fig1_example}. Given these unique distinctions against daily procedures, accurate interpretation of protocols requires explicit depiction of constraints (see \cref{fig:Fig1.5_program_examples} for example). 

\begin{figure*}[t!]
    \centering
    \includegraphics[width=\textwidth]{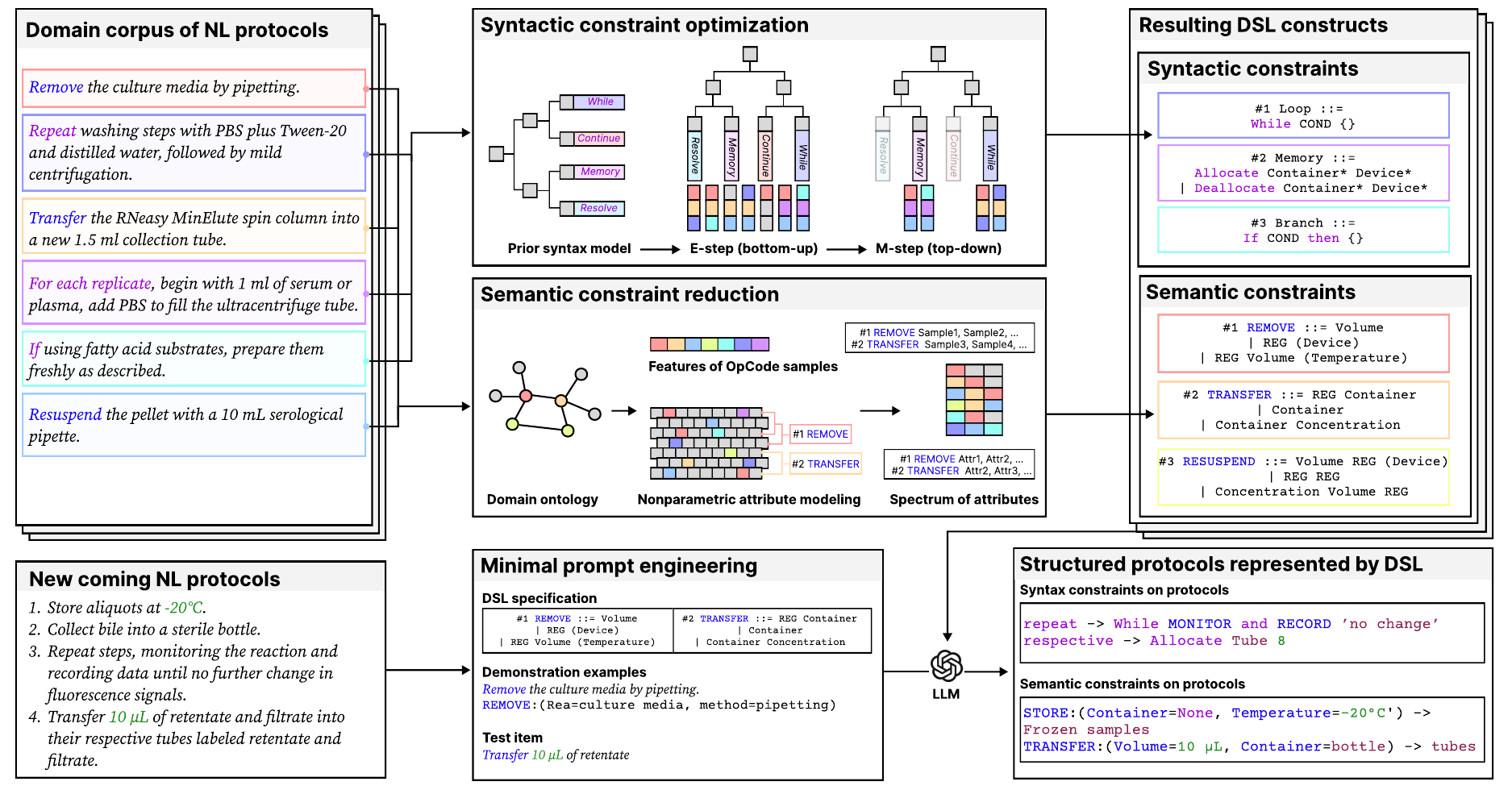}
    \caption{\textbf{The \ad framework and the resulting \ac{dsl}-based procedure constraints.} \textbf{(Top)} Given domain specified corpus, \ad conducts bidirectional syntax optimization and non-parametric semantic reduction, resulting in syntactic constraints and semantic constraints. \textbf{(Bottom)} A \ac{dsl}-based constraint takes novel procedures as input, handles the nonlinear syntax structures like \texttt{loop} by syntactic constraints, and handles the semantic errors like \texttt{missing} by semantic constraints.}
    \label{fig:Fig2_pipeline}
\end{figure*}

Intuitively, structural representation imposes constraints on the processing of protocols. This is achieved through purely symbolic approaches that depict procedures as flow-graphs~\citep{momouchi1980control,zhang2012automatically,kiddon2015mise}, and neuro-symbolic hybrid methods that superimpose procedural structures onto neural networks~\citep{puig2018virtualhome,huang2021scallop,huang2022language,zhou2022hierarchical,zhang2023causal,brohan2023can}. Both strategies impose constraints on the interpretation of procedures, thereby reducing the incidence of superfluous operations. Symbolic constraints range from elementary grammars to general-purposed programming languages, with the capability of a constraint system being pivotal in refining a coarse interpretation space into a more precise one~\citep{tarski1946introduction,chomsky1957syntactic,hopcroft1996introduction,russell2010artificial}. 

In light of the context, what level of capability should we expect the constraint to possess? This discourse introduces \ac{dsl}, a category of symbolic systems endowed with the most potent constraints. \acp{dsl} are programming languages tailored to specific problem domains, encapsulating both \textbf{syntactic constraints} and \textbf{semantic constraints} inherent to those domains. For instance, BioCoder~\citep{ananthanarayanan2010biocoder}, developed by Microsoft, is a \ac{dsl} explicitly designed to constrain experimental protocols. On the syntactic level, the variable management mechanism inherited from C/C++ enables \acp{dsl} to monitor the lifecycle of each intermediate product, ensuring no omissions or duplications. On the semantic level, the precise definitions of actions, combinations of reagents' names and volumes, with subprocedures abstracted from domain-specific concepts, guaranteeing procedural execution consistency. 

Drawing inspiration from \acp{dsl}, can we design constraints for protocols in \ac{dsl} fashion? Hardly, due to the deterministic and substantial cost. Structural constraints necessitate custom design for particular domains, which is prohibitively expensive, given that these domains are highly specialized and often diverge significantly from the conventional purview of computer scientists~\citep{mernik2005and,fowler2010domain}. The development of a \ac{dsl} necessitates a comprehensive integration of in-depth domain knowledge. Furthermore, the designed \ac{dsl} must align with theoretical aspects of formal language design while also meeting the distinct requirements of the specialized domain. This necessitates a bidirectional alignment between computer scientists and domain experts, a process that is intrinsically case-by-case, implying that a \ac{dsl} developed for one domain is unlikely to be applicable or easily adaptable to another, thus limiting the scalability of \ac{dsl}-based constraints across various domains. 

In this study, our objective is to offer an initial proof-of-concept aimed at reducing the design cost of \ac{dsl}-based constraints for protocols. We propose a scalable framework, termed \ad\!\!, that facilitates the automated creation of \acp{dsl}. The framework approaches the task as a bidirectional optimization problem, where the design of a \ac{dsl} is abstracted from domain-specific corpora through a bottom-up process and concurrently derived by general programming language design principles in a top-down manner. This approach emulates the iterative dialogue between computer scientists and domain experts, progressively bridging the conceptual gap between their respective fields of expertise. The syntactic constraint should adequately define consecutive actions and their repetitions, interruption, concurrence, subcomponents, and reactants. Constructs of the semantics constraint need to accurately reflect the domain's concepts and the relations between them, without redundancy or incompleteness.

We utilize protocols from various domains within the experimental sciences --- namely \textit{Genetics}, \textit{Medical and Clinical Research}, \textit{Ecology and Environmental Research}, \textit{Bioengineering}, and \textit{Computational Biology} --- as the primary testing ground for our methodology, due to their inherent complexity, resource constraints, and susceptibility to errors. These domains exhibit significant disparities both in syntactic and semantic language features. Comprehensive experiments demonstrate that \ad is capable of generalizing \ac{dsl}-based constraints tailored to these diverse domains, upholding the integrity from both programming language design and domain expertise perspectives. We further demonstrate that syntactic and semantic constraints effectively work as an auxiliary module of \acp{llm} in the processing of unseen protocols, thereby suggesting a promising future for constraining protocols through a synergistic blend of programs and natural language.

The contributions of this work are threefold: (i) We introduce the \ad framework for automated design of \ac{dsl}-based constraints, which includes a bidirectional syntax optimization module and a non-parametric learning module for semantic reduction. (ii) We establish a systematic and end-to-end evaluation platform to assess the quality of the designed \acp{dsl}-based constraints, employing both quantitative and qualitative metrics. (iii) We showcase the efficacy of \ac{dsl}-based constraints in processing new coming protocols with syntactic complexity and semantic errors. 

\section{Constraints in protocols}

In this section, we scrutinize the requirements for precisely constraining protocols. The distinctive challenges of engaging with such systems stem from their complexity of procedures, limited in resources, and vulnerability to errors. 

\paragraph{Complexity of procedures}

The complexity of protocols arises from the multitude of action types necessary for conducting experiments, the extensive categories of reagents involved, the variety of containers and devices for operational implementation, and the broad array of additional conditions affecting action execution, such as duration, temperature, volume, lighting, and acidity. For instance, whereas each daily procedure on wikiHow involves about 14 steps, 27 actions, and 45 objects on average\footnote{The statistics are calculated from the WikiHow Dataset, with 230k procedures in total~\citep{koupaee2018wikihow}.}, this figure substantially increases in the context of the five experimental science domains, where each protocol possesses around 60 steps, 76 actions, and 180 objects on average, with an increment of 250\% to 350\%. This necessitates the utilization of more specialized data types --- such as \texttt{Operations}, \texttt{Reagents}, \texttt{Conditions} --- over general data types like integers, floats, characters, and strings. Consequently, procedural workflows may include non-linear elements like loops for repetition, branches for parallel options, and subprocedures for nested and reusable actions, requiring specialized control flow structures to depict these complex scenarios.

\paragraph{Limitation in resources}

The execution of protocols is limited by the availability of resources, including a finite stock of reagents, a limited number of containers, and a scarcity of critical devices, characteristic of medium-scale experimental science laboratories. This parallels the allocation of registers and memory in computing, where each action must account for the use of memory spaces. In experimental procedure execution, if a device is currently in use, any operations requiring that device must be deferred until it becomes available. This necessitates resource management strategies similar to those in computer programming, such as allocating resources for use and deallocating them post-use. When allocated, the resource becomes inaccessible, creating bottlenecks that introduce inefficiency. To mitigate this, operations that do not compete for the same resources may be executed in parallel.

\paragraph{Vulnerability to errors}

Execution of protocols cannot tolerate errors, mandating strict adherence to every operation. This makes the system exceptionally vulnerable, in contrast to the robustness of everyday procedural execution. In such a context, protocol execution should encompass: (i) syntactic constraints specifying how actions are sequenced to form valid execution; and (ii) semantic constraints verifying that the reactants and reaction conditions are correctly utilized.

\begin{figure*}[t!]
    \centering
    \includegraphics[width=\textwidth]{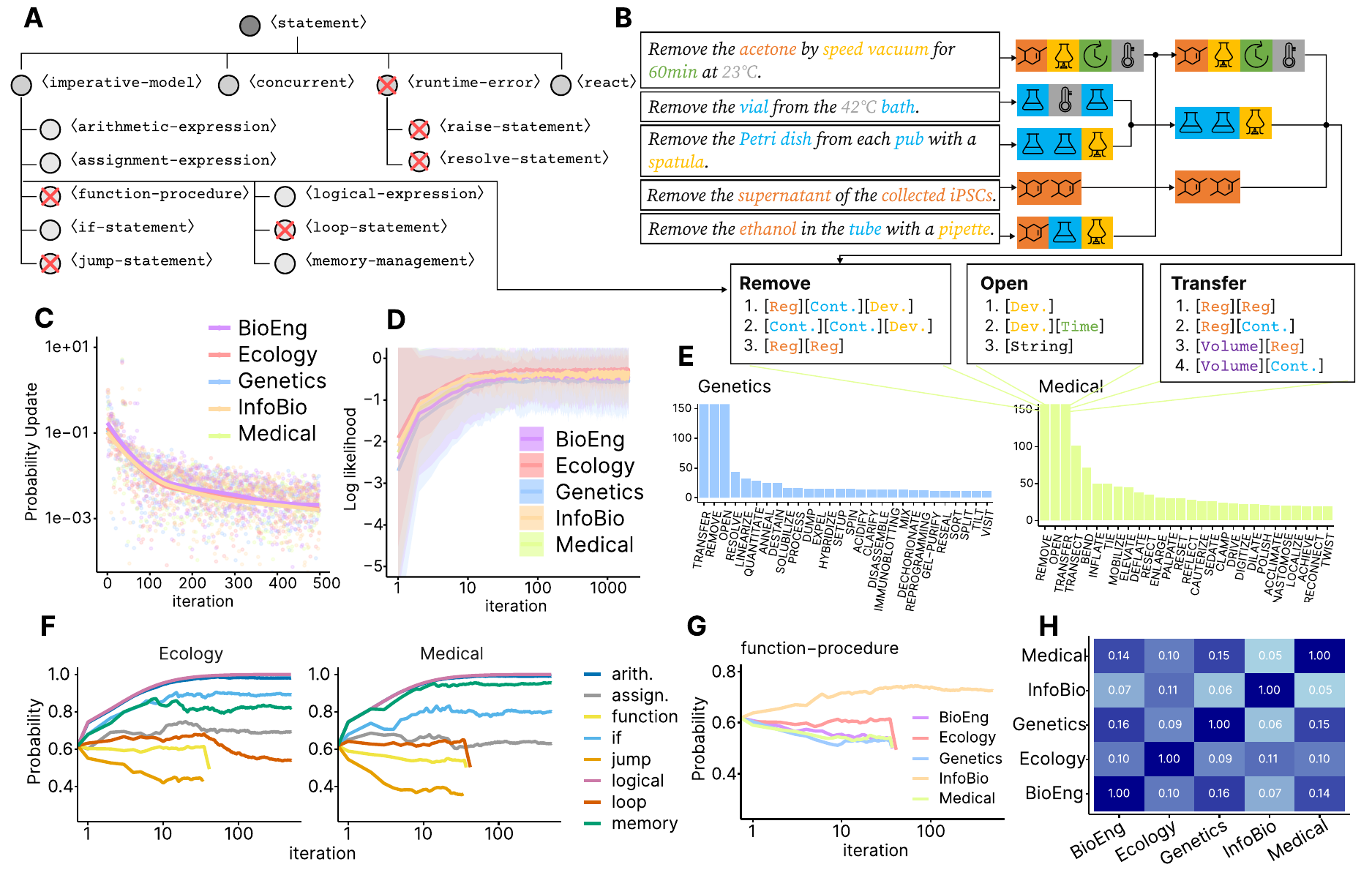}
    \caption{\textbf{Illustration on syntactic constraint optimization and semantics constraint reduction.} \textbf{(A)} Resulting syntactic constraints derived from the CFG prior model. \textbf{(B)} Resulting semantic constraints. \textbf{(C)} Convergence curve of syntactic constraint optimization. \textbf{(D)} Convergence curve of semantic constraint reduction. \textbf{(E)} Frequency profile of the semantic constraints of Genetics-DSL and Medical-DSL. \textbf{(F)} Acquisition of different syntactic constraints on Ecology and Medical domain corpora. \textbf{(G)} Syntactic constraint \texttt{Function-procedure} is differently acquired by the five distinct domains. \textbf{(H)} Confusion matrix indicating the overlapped semantic constraints between the five distinct domains.}
    \label{fig:Fig3_algorithm}
\end{figure*}

\section{\ad framework}

This section describes our problem formulation and solutions for the \ad framework.

\subsection{Problem formulation}

\paragraph{Input} 

The system input consists of natural language descriptions of protocols, namely a domain-specific corpus. These descriptions encompass procedural knowledge of execution orders and ingredient knowledge of reagents, containers, and devices involved in the experiments. For a specific domain, the input corpus $\mathcal{C}=\{\mathbf{c}_1, \mathbf{c}_2, \dots, \mathbf{c}_N\}$ includes $N$ protocols.

\paragraph{Output}

The desired output is a \ac{dsl} $\{\mathcal{S}, \Lambda\}$, incorporating sets of constructs that define both syntactic and semantic constraints specific to the domain~\citep{fowler2010domain}, while retaining the abstract qualities of a programming language. The set $\mathcal{S}=\{\varphi_1,\dots,\varphi_K\}$ comprises $K$ atomic syntactic constraints with production rules, such as control structures \texttt{Loop}, \texttt{Parallel}, \texttt{Jump}, and \texttt{Split}. The set $\Lambda=\{t_1,\dots,t_L\}$ embodies $L$ atomic semantic constraints, such as operations \texttt{Add}, \texttt{Remove}, \texttt{Incubate}, and \texttt{Place}.

\subsection{Syntactic constraint optimization}

\paragraph{Key insight}

Leveraging existing knowledge on programming language design, our method utilizes a bidirectional optimization strategy to formulate the syntax of the target \acp{dsl} based on the prerequisites of the domain corpora. The algorithm employs an \ac{em} framework, where the E-Step abstracts syntax from domain corpora and the M-Step derives syntax from programming language principles.

\paragraph{Modeling}

The algorithm models latent syntactic constraint assignments $\mathcal{Z}=\{z_1,\dots,z_N\}$ for each protocol $c_i$. A filter set $\Theta=\{\theta_1,\dots,\theta_{K'}\}$, where $K'>K$, is designed to determine if a segment of procedural text aligns with the logic of any atomic syntactic constraint, coming with the belief function $p(\Theta|\mathcal{S})$. The observational likelihood is computed as $p(\mathcal{C}|\mathcal{Z},\Theta)=\prod^N_{i=1}p(x_i|z_i,\theta_{z_i})$. Hence, the overall joint probability of the model is given by:
\begin{equation}
    p(\mathcal{C},\mathcal{Z},\Theta|\mathcal{S})=p(\mathcal{C}|\mathcal{Z},\Theta)p(Z|\mathcal{S})p(\Theta|\mathcal{S}).
\end{equation}

\paragraph{Syntax prior}

Programming language designers leverage a general set of syntactic production rules as the prior $p(Z|\mathcal{S})$ for syntax specification. We initialize the set $\mathcal{S}_0$ with a \ac{cfg}~\citep{hopcroft1996introduction} (see \cref{fig:Fig3_algorithm}A). Furthermore, we construct the prior belief function $p(\Theta|\mathcal{S})$ with a series of sliding-window-based filters $f: \mathcal{C} \mapsto \mathbb{R}$, which gives a relaxed lower bound for predicting the existence of an atomic syntactic constraint. Please refer to \cref{sec:implement-detail} for details.

\paragraph{E-Step}

In each E-Step, we obtain the posterior of latent variables $p(\mathcal{Z}|\mathcal{C},\Theta,\mathcal{S})$ applying Bayes' theorem, which is implemented by scanning the filters over domain corpus. To note, as the spaces of prior and observation are not intractably large, we simply employ the naive version of E-Step without variational approximations.

\paragraph{M-Step}

In each M-Step, we first maximize the atomic syntactic constraints $\mathcal{S}$ by maximizing:
\begin{equation}
    \mathcal{Q}(\Hat{\Theta},\Theta)=\mathbb{E}_{\mathcal{Z}|\mathcal{C},\Theta}\big[\log p(\mathcal{C},\mathcal{Z},\Hat{\Theta}|\mathcal{S})\big],
\end{equation}
where $\Hat{\Theta}$ is the updated $\Theta$, resulting in the structural change of $\mathcal{S}_0$. These two steps alternate iteratively until convergence (see \cref{fig:Fig3_algorithm}C), ensuring the syntactic constraints are aligned with the domain.

\subsection{Semantic constraint reduction}

\paragraph{Key insight}

Following the adaptation of syntactic constraints to the target domain, the semantic reduction phase focuses on distilling fine-grained semantic constraints. This stage addresses the absence of the domain-specific prior knowledge on semantics by employing a non-parametric approach, \ie, \ac{dpmm}, which allows for the flexible identification of semantic concepts and their relationships inherent within the protocols (see \cref{fig:Fig3_algorithm}B).

\paragraph{Modeling}

We transform the protocols into a vectorized feature space, $\mathcal{X}=\{\mathbf{x}_1,\dots,\mathbf{x}_N\}$, with each $\mathbf{x}_i$ encoding operation patterns in a one-hot vector format (refer to \cref{sec:Apdx-feature-vector}). The objective is to model latent semantic constraint assignments, $\mathcal{W}=\{w_1,\dots,w_N\}$, where each $w_i$ is an assignment of atomic semantic constraint on $\mathbf{x}_i$. To note, according to the definition, the size of the semantics set $\Lambda$ is not fixed and grows with the data. 

\paragraph{Iteration}

Utilizing a \ac{dpmm} facilitates the non-parametric spectral clustering of these feature vectors into groups of unique atomic semantic constraints. Each iteration in the \ac{dpmm} process involves Gibbs Sampling for estimating the posterior of model's parameters (see \cref{fig:Fig3_algorithm}D). This clustering not only identifies distinct semantic operations but also adapts to the complexity and variability of semantic patterns across different domain corpora. Refer to \cref{sec:Apdx-dpmm} for details.  

\subsection{Discussion}\label{subsec:algo-discuss}

We identify the commonalities and distinctions between the \acp{dsl} designed by \ad corresponding to the five experimental science domains (see \cref{fig:Fig3_algorithm}H). We find that the five domains share a majority of syntactic constraints, such as \texttt{Memory} and \texttt{Branch} (see \cref{fig:Fig3_algorithm}F), while differing in other constraints, such as \texttt{Function-procedure} and \texttt{Loop} (see \cref{fig:Fig3_algorithm}G). This implies that the domain specificity as an experimental protocol dominates that of subject, regarding the syntax.

Different from syntactic constraints, the five domains vary significantly on semantic constraints, including fine-grained concepts such as operations, reagents, and conditions (see \cref{fig:Fig3_algorithm}E). This implies that the domain specificity of knowledge ontology is dominant regarding the semantics.

\begin{figure*}[t!]
    \centering
    \includegraphics[width=\textwidth]{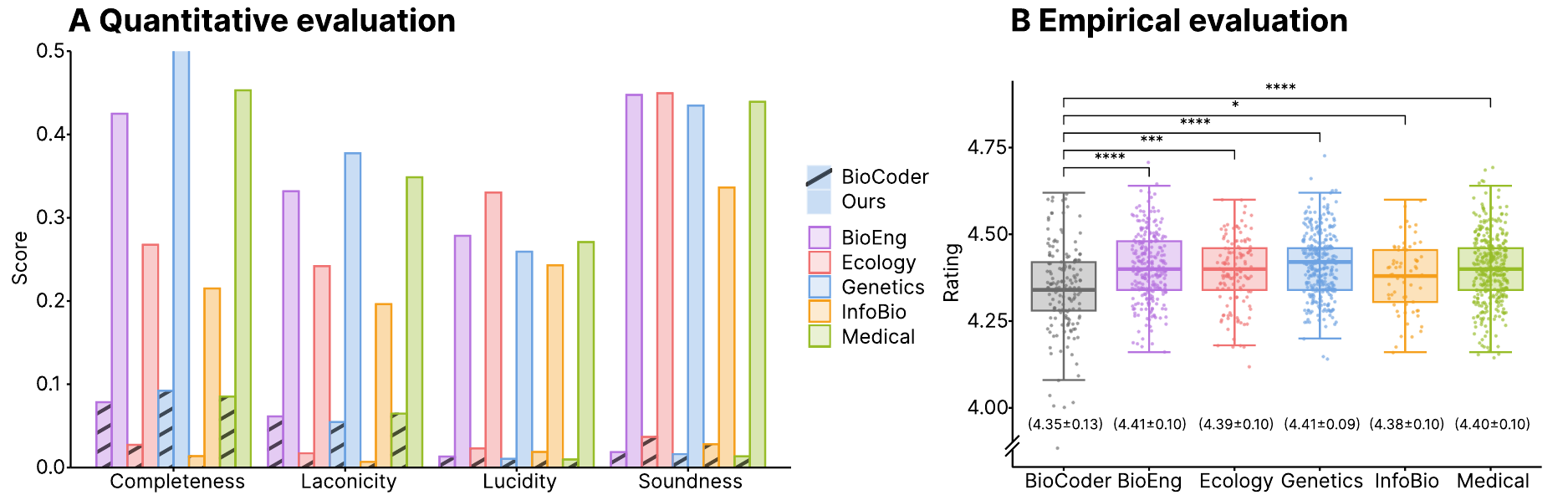}
    \caption{\textbf{Constraint design results on 5 experimental science domains.} \textbf{(A)} Quantitative evaluation on the 5 \ac{dsl}-based constraints and the BioCoder baseline. \textbf{(B)} Empirical evaluation on the 5 \ac{dsl}-based constraints and the BioCoder baseline.}
    \label{fig:Fig4_res_design}
\end{figure*}

\section{Constraint design assessment}\label{sec:res}

In this section, we evaluate the quality of the \acp{dsl} automatically designed by our \ad framework across the five domains. We first discuss the principles for the assessment, and then conduct quantitative and qualitative analysis accordingly. 

\subsection{Domain corpora}

We compile a dataset of 16,194 experimental protocols across five domains: Genetics (8794 protocols), Medical and Clinical Research (7351, ``Medical'' for abbreviation), Ecology and Environmental Research (812, ``Ecology''), Bioengineering (3597, ``BioEng''), and Computational Biology (348, ``InfoBio''), with minimal overlap between them. This diverse dataset, essential for testing our methodology, reflects the unique syntactic and semantic requirements of each domain. Please refer to \cref{sec:s-data-corpora} for details on data collection.

\subsection{What makes a good \ac{dsl}?}

We leverage a systematic approach to gain insight into what constitutes the appropriate objectives of \acp{dsl}, employing the frameworks introduced by~\citet{guizzardi2013ontology} for quantitative and~\citet{karsai2009design} for empirical assessment. 

\paragraph{Quantitative evaluation} 

We check the mapping from ontology elements in the reference model, \ie, concepts and relations in the domain corpus, to \ac{dsl} constructs of constraints, focusing on four criteria: \emph{soundness}, \emph{lucidity}, \emph{completeness}, and \emph{laconicity}. An ideal \ac{dsl} should guarantee that (i) all ontology elements are mapped to the construct set for \textbf{soundness}; (ii) every ontology element is mapped to only one construct for \textbf{lucidity}; (iii) the construct set does not possess any redundancy beyond the ontology elements for \textbf{completeness}; (iv) every construct refers to only one ontology element for \textbf{laconicity}. 

\paragraph{Empirical evaluation} 

We take the design guidelines of \ac{dsl} as questions for assessing the resulting \acp{dsl} by \ad, from a user-centric perspective. These questions range from the objective consistency for machine interpretation to the subjective complexity for user understanding. Specifically, the \ac{dsl} should maintain (i) \textbf{simplicity}, \ie, being easy to understand both syntactically and semantically; (ii) \textbf{clarity}, \ie, pruning unnecessary space for generalization; and (iii) \textbf{conciseness}, \ie, avoiding redundant concepts and relations. Please refer to \cref{sec:Apdx-empirical-eval} for details.

\subsection{Quantitative evaluation}

\paragraph{Method} 

To quantitatively evaluate the effectiveness of \ac{dsl}-based constraints in protocol representation, we leverage maximum-recall domain-specified ontology knowledge extracted from domain-specific corpora. Such ontology, represented as a graph consisting of concept-relation triplets, serves as the groundtruth for our assessment on the \acp{dsl} designed by \ad\!\!. 

\paragraph{Results} 

The analysis on the \acp{dsl} designed by \ad, comparing with BioCoder across several domains, is detailed in \cref{fig:Fig4_res_design}A. Specifically, our \acp{dsl} adeptly model 43.47\% of the concepts with precise constraints, \ie, soundness, of which 25.93\% showcase a direct one-to-one correspondence, \ie, lucidity. Furthermore, the completeness metric reveals that 50.51\% of DSL constraints correspond with specific domain concepts, and 37.74\% of constraints are uniquely aligned with a single domain concept, \ie, laconicity. In contrast, BioCoder demonstrates significantly lower performance metrics (lucidity: 1.05\%, soundness: 1.61\%, laconicity: 5.46\%, completeness: 9.22\%). This trend of a 5-to-20-fold improvement in the resulting \acp{dsl} over BioCoder on each metric is consistent across the five domains, as highlighted in our results.

\paragraph{Discussion} 

The quantitative evaluation underscores the superiority of the \acp{dsl} designed through \ad over BioCoder, the established \ac{dsl} hand-crafted by domain experts, in terms of four quantitative metrics. Despite the inherent specificity of the protocols, this assessment highlights that a compact subset of constraints can be precisely defined and extracted. These results validate our framework's ability to model constraints that more accurately and effectively encapsulate domain-specified procedural knowledge. 

\subsection{Empirical evaluation}

\paragraph{Method}

We institute an automatic evaluation framework leveraging a state-of-the-art \ac{llm}, \eg, GPT-4, augmented with a novel \ac{qa} designed to simulate the analytical capabilities of human domain experts. This \ac{qa} is engineered using an index vector embedding technique to integrate domain-specific corpora from the five domains into the \ac{llm}. To facilitate nuanced and domain-aware responses, we employed advanced prompt-engineering strategies, including the \ac{cot} technique~\citep{wei2022chain}, enabling the \ac{qa} to generate assessments that closely mirror those of a domain expert. The system's empirical evaluation was predicated on its performance across a set of 50 meticulously designed questions, as delineated in \cref{sec:Apdx-empirical-eval}.

\paragraph{Meta-evaluation}

Prior to deploying this automated evaluation mechanism across our \acp{dsl}, we conduct a preliminary meta-evaluation to ascertain the congruence between the \ac{qa}'s assessments and those of human domain experts, following the routine of automatic evaluation~\citep{schuff2023human}. This process involves a comparative analysis on four subsets extracted from our \acp{dsl} and BioCoder, with both human domain experts $(N=3)$ and the \ac{qa} providing ratings. We observe no significant evidence supporting that human experts and the \ac{qa} rate differently ($t(99)=-1.282$, $\mu_d\neq 0$, $p=.202$), validating the application of our automated evaluation framework in assessing the resulting \acp{dsl}.

\paragraph{Results}

In the ensuing phase, the \ac{qa} appraises the five \acp{dsl}, yielding ratings and confidence scores. A subsequent analysis employing paired samples T-tests (see \cref{fig:Fig4_res_design}B; $t(49)=3.487$, $\mu_d\leq 0$, $p < .005$) substantiated that our resulting \acp{dsl} exhibit quality metrics on par with those of BioCoder --- the benchmark \ac{dsl} meticulously crafted using C/C++.

\paragraph{Discussion}

The empirical evaluation not only underscores the viability of our \ad framework in automating the design of domain-specific languages, but also highlights its potential to match or even surpass the quality of manually engineered solutions like BioCoder. It is primarily attributed to the design decisions to optimize a broad language feature set for compact syntactic constraints; and to tailor operations for the domain for precise semantic constraints. These strategies establish a clean and compact \ac{dsl}-based constraint that adheres to domain conventions.

\begin{figure*}[t!]
    \centering
    \includegraphics[width=\textwidth]{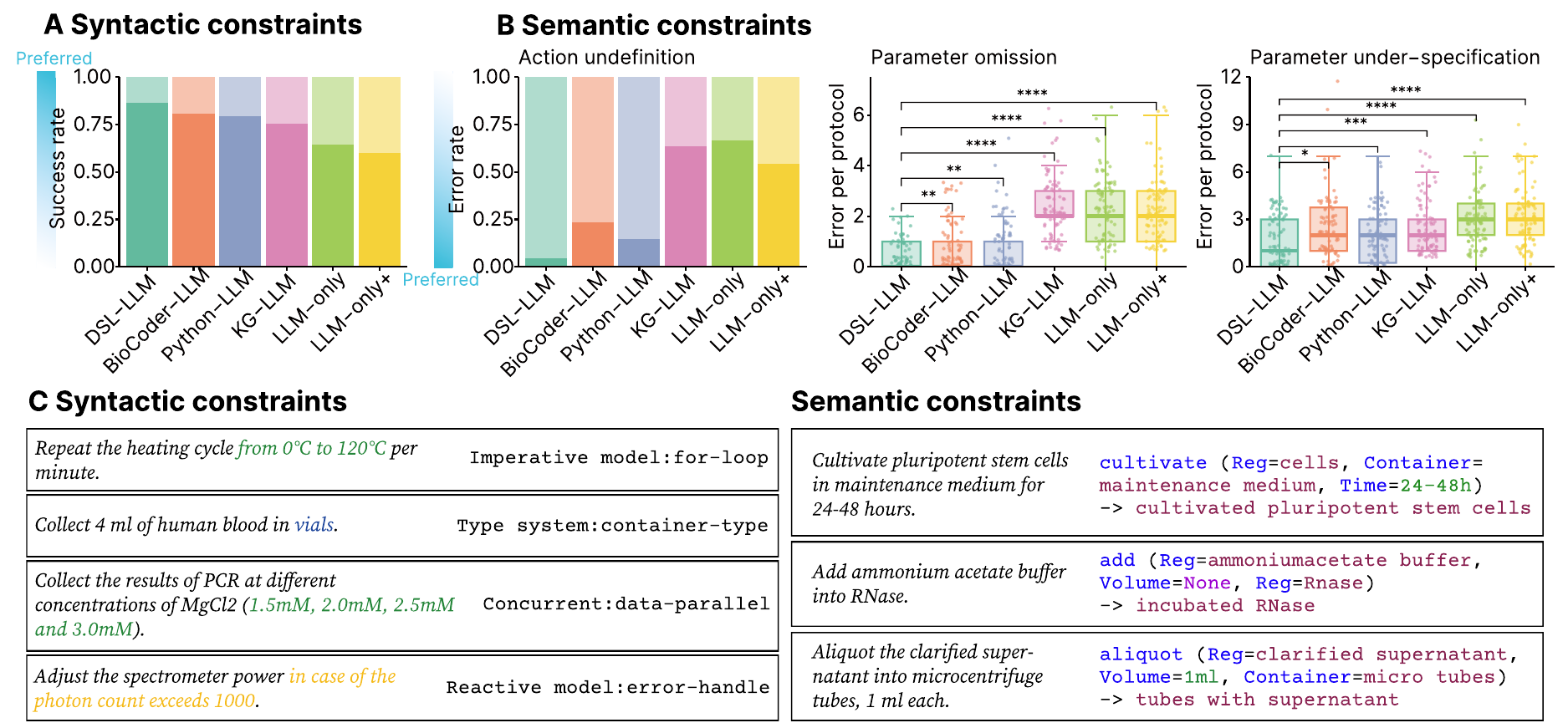}
    \caption{\textbf{Constraint utility assessment results on the five experimental science domains between \ac{dsl}-based constraint and alternative models.} \textbf{(A)} Results on syntactic constraint utility assessment \textbf{(higher is preferred)}. \textbf{(B)} Results on semantic constraint utility assessment \textbf{(lower is preferred)}. \textbf{(C)} New coming protocols represented with syntactic or semantic constraints.}
    \label{fig:Fig5_res_utility}
\end{figure*}

\section{Constraint utility assessment}

In this section, we evaluate the utility of leveraging \ac{dsl}-based constraints for the representation of new coming experimental protocols across the five domains, considering both syntax and semantics. Please refer to \cref{tab:showcases} for a demonstration.

\subsection{Materials}

Our dataset includes 186 new coming protocols, meticulously collected from recent experiments by domain experts across the specified domains. These protocols, which have been validated for accuracy, serve as a testbed for evaluating the performance of \ac{dsl}-based constraints on \textbf{unseen} data during the design phase. We identify the challenges with four syntactic features --- Imperative control flows, Type system, Concurrent, and Reactive model. There are also three semantic errors --- Action undefinition, Parameter omission, and Parameter under-specification.

For syntactic constraints, we determine the success rate by calculating the proportion of samples where the target challenging syntactic features are accurately constrained. Similarly, for semantic constraints, the success rate is assessed by the proportion of samples where the \ac{dsl} successfully identifies and resolves semantic errors. For fair comparison, we transform the varied outputs from different approaches to a unified JSON-style representation. For strings referring to names of instances, we relax the exact-match criteria to similarity-based score~\citep{papineni2002bleu}.

\subsection{Methods}

We employ a multi-dimensional approach to assess the utility of \ac{dsl}-based constraints, comparing them with several alternative methods. This maintains the integrity of an end-to-end workflow with minimal adjustments to the \ac{llm}.

\paragraph{Our \ac{dsl}-based constraint \& \ac{llm}}

Incorporating \ac{dsl} as an external constraint involves using it as an interpreter for the programs generated by the \ac{llm}. Following protocols suggested by recent research~\citep{gao2023pal,zhang2023causal}, we prompt the \ac{llm} with \ac{dsl} syntax grammar and semantics operation set with running examples, directing it to translate procedural texts into corresponding programs. These programs are then verified and potentially corrected using the \ac{dsl}'s syntactic and semantic constraints. This approach minimizes external dependencies, striving for a seamless plug-and-play integration between \acp{dsl} and \acp{llm}. We utilize GPT-3.5 as the backbone model, resulting in \texttt{DSL-LLM}.

\paragraph{Programming language \& \ac{llm}}

As comparative baselines, we explore the use of another \ac{dsl}, BioCoder, and a general-purpose programming language, Python, with GPT-3.5 as the base model. These methodologies are denoted as \texttt{BioCoder-LLM} and \texttt{Python-LLM}, respectively. The main variance lies in the adoption of one-shot generation for fair comparison, given GPT's pre-existing familiarity with both BioCoder and Python's coding paradigms.

\paragraph{Structural representation \& \ac{llm}}

We also examine the effectiveness of elementary structural knowledge representation, specifically \ac{kg}, as a simpler alternative. This method, \texttt{KG-LLM}, leverages entity-relation extraction techniques for knowledge structuring.

\paragraph{\ac{llm} only}

Finally, we assess the capability of naive \acp{llm} operating without any structural constraints, particularly using GPT-3.5 and GPT-4, to gauge the impact of \ac{llm} advancements alone, named after \texttt{LLM-only} and \texttt{LLM-only+}.

\subsection{Results}

\paragraph{Syntactic constraints}

Our investigation into syntactic constraints shows that \texttt{DSL-LLM} surpasses all alternative approaches in terms of success rate across the five domains (see \cref{fig:Fig5_res_utility}A). In the analysis of syntactic constraints across five domains, \texttt{DSL-LLM} achieves a success rate of 93.5\%, which significantly outperforms the counterparts without programming language representation ($\chi^2(1)=3.979, p<.05$), and also outperforms the counterparts with general syntactic constraints. This substantiates the efficacy of these domain-specific syntactic constraints in meticulously constraining non-trivial protocols.

\paragraph{Semantic constraints}

The assessment of semantic constraints reveals that \texttt{DSL-LLM} outperforms both the strong and weak baselines (see \cref{fig:Fig5_res_utility}B) regarding the three types of errors. In the context of Genetics domain, \texttt{DSL-LLM} yields a success rate of 93.7\% in addressing semantic errors, significantly outperforms alternative methods, as indicated by statistical evidence ($\chi^2(1)=8.378, p < .005$ in action undefinition; $t(185)=-3.215, \mu_d < 0, p <. 005$ in parameter omission; and $t(185)=-2.164, \mu_d < 0, p <. 05$ in parameter under-specification). These underscore the domain-specific semantic constraints' capability in enhancing accuracy of protocol representation, which is crucial for successful experiments.

\begin{table*}[t]
    \centering 
    \caption{\textbf{Showcases of protocol representations with all approaches.} (OP: original protocol; GT: ground truth; L+: \texttt{LLM-only+}; each abbreviation consists of the two initials from both sides of the dash in each approach name)}
    \fontsize{7}{8}\selectfont
    \rowcolors{1}{gray!20}{white}
    \renewcommand{\arraystretch}{1.25}
    \begin{tabularx}{\linewidth}{c|X|X|>{\arraybackslash}X}
    \toprule
    OP & \textbf{RNA Determination:} \emph{Add ammonium acetate buffer and RNaseT2, then incubate.} & \textbf{Bile Processing:} \emph{Use a centrifuge tube to spin at 3000g for 10 min at 4°C.} & \textbf{Bile Collection:} \emph{Collect bile into a sterile collection bottle.}\\
    
    GT & \cdfont{\cdfun{ADD}: [[\cdfun{Reg}: \cdstr{ammonium acetate buffer}], [\cdfun{Container}: \cdkey{None}], [\cdfun{Volume}: \cdkey{None}], [\cdfun{Reg}: \cdstr{RNaseT2}]] -> \cdstr{incubated RNaseT2}} & \cdfont{\cdfun{SPIN}: [[\cdfun{Force}: \cdcon{3000g}], [\cdfun{Time}: \cdcon{10min}], [\cdfun{Temperature}: \cdcon{4°C}], [\cdfun{Container}: \cdstr{centrifuge tube}]] -> \cdstr{centrifuged sample}} & \cdfont{\cdfun{COLLECT}: [[\cdfun{Reg}: \cdstr{bile}], [\cdfun{Volume}: \cdkey{None}], [\cdfun{Container}: \cdstr{sterile collection bottle}]] -> \cdstr{bile sample}}\\
    
    DL & \cdfont{\cdfun{ADD}: [[\cdfun{Reg}: \cdstr{ammonium acetate buffer}],  [\cdfun{Container}: \cdkey{None}], [\cdfun{Volume}: \cdkey{None}], [\cdfun{Reg}: \cdstr{RNaseT2}]] -> \cdstr{incubated RNaseT2}} & \cdfont{\cdfun{SPIN}: [[\cdfun{Force}: \cdcon{3000g}], [\cdfun{Time}: \cdcon{10min}], [\cdfun{Container}: \cdstr{centrifuge tube}], [\cdfun{Temperature}: \cdcon{4°C}]] -> \cdstr{centrifuged sample}} & \cdfont{\cdfun{COLLECT}: [[\cdfun{Reg}: \cdstr{bile}], [\cdfun{Volume}: \cdkey{None}], [\cdfun{Container}: \cdstr{sterile collection bottle}]] -> \cdstr{bile sample}}\\
    
    BL & \cdfont{\cdfun{ADD}: [[\cdfun{Reg}: \cdstr{ammonium acetate buffer}], [\cdfun{Volume}: \cdkey{None}], [\cdfun{Reg}: \cdstr{RNaseT2}]] -> \cdstr{incubated RNaseT2}} & \cdfont{\cdfun{SPIN}: [[\cdfun{Container}: tube], [\cdfun{Force}: \cdcon{3000g}], [\cdfun{Time}: \cdcon{10min}], [\cdfun{Temperature}: \cdcon{4°C}]] ->} & \cdfont{\cdfun{COLLECT}: [[\cdfun{Container}: \cdstr{sterile collection bottle}]] -> centrifuged pellet}\\
    
    PL & \cdfont{\cdfun{ADD}: [[\cdfun{Reg}: \cdstr{ammonium acetate buffer}], [\cdfun{Volume}: \cdkey{None}], [\cdfun{Reg}: \cdstr{RNaseT2}]] -> \cdstr{incubated RNaseT2}} & \cdfont{\cdfun{SPIN}: [[\cdfun{Force}: \cdcon{3000g}],  [\cdfun{Time}: \cdcon{10min}], [\cdfun{Temperature}: \cdcon{4°C}]] -> \cdstr{centrifuged sample}} & \cdfont{CENTRIFUGE: [[\cdfun{Container}: bottle]] -> centrifuged bile} \\
    
    KL & \cdfont{: [] ->} & \cdfont{STEP1: [[\cdfun{Force}: \cdcon{3000g}], [\cdfun{Time}: \cdcon{10min}], [\cdfun{Temperature}: \cdcon{4°C}]] ->} & \cdfont{: [] ->} \\
    
    LO & \cdfont{: [] ->} & \cdfont{STEP: [[Device: centrifuge],  [\cdfun{Force}: \cdcon{3000g}], [\cdfun{Time}: \cdcon{10min}], [\cdfun{Temperature}: \cdcon{4°C}]] ->} & \cdfont{CENTRIFUGE: [[\cdfun{Container}: \cdstr{sterile collection bottle}]] -> centrifuged bile}\\
    
    L+ & \cdfont{\cdfun{ADD}: [[\cdfun{Reg}: \cdstr{ammonium acetate buffer}], [\cdfun{Reg}: \cdstr{RNaseT2}]] -> \cdstr{incubated RNaseT2}} & \cdfont{\cdfun{SPIN}: [[Device: tube],  [\cdfun{Time}: \cdcon{10min}], [\cdfun{Force}: \cdcon{3000g}], [\cdfun{Temperature}: \cdcon{4°C}]] -> sample} & \cdfont{CENTRIFUGE: [[\cdfun{Reg}: \cdstr{bile}], [\cdfun{Container}: \cdstr{sterile collection bottle}]] -> collected bile}\\
    \bottomrule
    \end{tabularx}
    \label{tab:showcases}
\end{table*}

\subsection{Discussion}

The performance of alternative approaches basically aligns with our expectation (see \cref{fig:Fig5_res_utility}C). Among them, \texttt{DSL-LLM} demonstrates the highest performance in addressing both syntactic and semantic constraints. While \texttt{BioCoder-LLM} and \texttt{Python-LLM}, the strong baselines, are outperformed by \texttt{DSL-LLM} in most times, they still exhibit substantial advantages over other baselines. This phenomenon can be attributed to the merits shared by programming languages, such as the ability to represent structural knowledge at various levels of granularity. Although not as effective as semantics constraints, which exactly define the structures and legal ranges for the space of operations, reagents, and conditions, this type of representation still constrains the potential search space to some extent. On the other hand, the flattened structural representation of \acp{kg} cannot provide the same level of expression capacity for \texttt{KG-LLM} as programming languages. Despite the expectation that GPT-4 would be much more capable than GPT-3.5, the performances of \texttt{LLM-only} and \texttt{LLM-only+} are comparable, suggesting that a pure text representation may not be suitable for processing complicated procedures like protocols. The relative success of \texttt{DSL-LLM} indicates the potential of \acp{dsl} as external constraints for \acp{llm}. 

\section{General discussions}

In this work, we present the \ad framework as a proof-of-concept to facilitate the automation of designing \ac{dsl}-based constraints across various domains. Through both quantitative and qualitative evaluations of the \acp{dsl} designed by \ad in five distinct domains, we demonstrate its capability as an auxiliary module for \ac{llm}.

\paragraph{Rationale behind \ac{dsl}}

The decision to leverage \ac{dsl} for constraint representation is rooted in several considerations. Primarily, \acp{dsl} adeptly capture domain-specified syntactic and semantic constraints, aligning well with the hardness of representing protocols with complicated control flows and operations. Moreover, \acp{dsl} leverage a deterministic verification mechanism derived from general-purpose programming languages, offering a robust means of imposing constraints on the inherently nondeterministic outputs of \acp{llm}. Additionally, \acp{dsl} are user-friendly to both humans and machines, maintaining a minimal set of language features that facilitate ease of adoption.

\paragraph{No universal constraint}
 
It is unrealistic to expect a one-size-fits-all protocol constraint applicable across varied domains. For a constraint system to accurately delineate the execution space of a particular domain, it must encompass domain-specific syntax and semantics. Though it is possible to devise a comprehensive set of constraints that covers the requirements of all conceivable domains, such an approach would yield a constraint system of prohibitive complexity, rendering it impractical for end-users. Conversely, simplifying this universal constraint to enhance user-friendliness inevitably compromises its capability, leading to the expressivity-complexity dilemma~\citep{abelson1996structure}. Rather than seeking an elusive generality, focusing on domain-specific constraint development and striving for the automation of this process may offer a pragmatic way to circumvent this dilemma.

\section*{Ethics statement}

\paragraph{Human participants}

The meta-evaluation and data annotation included in this work has been approved by the \ac{irb} of Peking University. We have been committed to upholding the highest ethical standards in conducting this study and ensuring the protection of the rights and welfare of all participants. Every domain expert was paid on a wage of \$22.5 per hour for participating in the meta-evaluation and data annotation. We have obtained informed consent from all participants, including clear and comprehensive information about the purpose of the study, the procedures involved, the risks and benefits, and the right to withdraw at any time without penalty. Participants were also assured of the confidentiality of their information. Any personal data collected, including name, age, gender, institution, and education background, was handled in accordance with applicable laws and regulations. 

\paragraph{Corpora collection}

We carefully ensure that all experimental protocols incorporated into our corpora strictly adhere to open access policies, governed by the Creative Commons license. This approach guarantees full compliance with copyright and intellectual property laws, eliminating any potential infringement or unauthorized use of protected materials. By exclusively utilizing resources that are freely available and legally distributable, we uphold the highest standards of ethical conduct in research, fostering an environment of transparency and respect for the intellectual property rights of others. This commitment ensures that our work not only advances the frontiers of knowledge but does so in a manner that is both legally sound and ethically responsible.

\section*{Limitations}\label{sec:lim}

As a proof-of-concept work, the design and evaluation of \ad come with limitations, leading to further investigations: 
\begin{itemize}[noitemsep,nolistsep,topsep=0pt,leftmargin=*]
    \item We majorly consider the imperative programming model as the prior model for \ac{dsl} design in the work. This raises the question of whether incorporating alternative programming paradigms, such as functional and object-oriented models, could enhance the representation of complex entities within protocols, particularly the properties of reagents.
    \item \ad's current framework outputs only syntactic and semantic constraints, lacking an explicit production system. This leads to the question of whether it is feasible to augment the \ad framework to autonomously generate both a production system and a constraint system, leveraging domain-specific corpora and pre-existing knowledge on programming languages. Such an enhancement could significantly improve \ac{dsl}'s potential on planning.
    \item To ensure a fair comparison and to underscore the plug-in capability of the developed \acp{dsl}, only minimal prompt engineering is applied in protocol processing. This posits the potential for developing tools that could more effectively intertwine \acp{dsl} with \acp{llm}.
    \item Notably, the DSL-InfoBio's quantitative evaluation outcomes are markedly inferior to those of its \ac{dsl} counterparts. Considering the notably smaller size of the InfoBio domain corpus relative to other domain-specific corpora, this prompts an investigation into the potential correlation between the scale of a domain corpus and the quality of the resulting \ac{dsl}. 
\end{itemize}
With many questions unanswered, we hope to explore more on automated design of \ac{dsl}-based constraints for procedural representation.

\section*{Reproducibility}

The project page with supplementary files for reproducing the results of this paper is available at \url{https://autodsl.org/procedure/papers/acl24shi.html}.

\section*{Acknowledgements}
This work was supported by the National Natural Science Foundation of China under Grant 91948302. Part of the authors are visiting students at Peking University during this work. In particular, Z. Bi is visiting from Huazhong University of Science and Technology, X. Wei is visiting from Beijing Jiaotong University. The authors would like to appreciate these institutes, and also Prof. Wenjuan Han for her support.

\bibliographystyle{acl_natbib}
\bibliography{references}

\begin{thebibliography}{42}
\expandafter\ifx\csname natexlab\endcsname\relax\def\natexlab#1{#1}\fi

\bibitem[{Abelson and Sussman(1996)}]{abelson1996structure}
Harold Abelson and Gerald~Jay Sussman. 1996.
\newblock \emph{Structure and interpretation of computer programs}.
\newblock The MIT Press.

\bibitem[{Ananthanarayanan and Thies(2010)}]{ananthanarayanan2010biocoder}
Vaishnavi Ananthanarayanan and William Thies. 2010.
\newblock Biocoder: A programming language for standardizing and automating biology protocols.
\newblock \emph{Journal of Biological Engineering}, 4(1):1--13.

\bibitem[{Baker(2016)}]{baker20161}
Monya Baker. 2016.
\newblock 1,500 scientists lift the lid on reproducibility.
\newblock \emph{Nature}, 533(7604).

\bibitem[{Brohan et~al.(2023)Brohan, Chebotar, Finn, Hausman, Herzog, Ho, Ibarz, Irpan, Jang, Julian et~al.}]{brohan2023can}
Anthony Brohan, Yevgen Chebotar, Chelsea Finn, Karol Hausman, Alexander Herzog, Daniel Ho, Julian Ibarz, Alex Irpan, Eric Jang, Ryan Julian, et~al. 2023.
\newblock Do as i can, not as i say: Grounding language in robotic affordances.
\newblock In \emph{Conference on Robot Learning}.

\bibitem[{Brown et~al.(2020)Brown, Mann, Ryder, Subbiah, Kaplan, Dhariwal, Neelakantan, Shyam, Sastry, Askell et~al.}]{brown2020language}
Tom Brown, Benjamin Mann, Nick Ryder, Melanie Subbiah, Jared~D Kaplan, Prafulla Dhariwal, Arvind Neelakantan, Pranav Shyam, Girish Sastry, Amanda Askell, et~al. 2020.
\newblock Language models are few-shot learners.
\newblock In \emph{Advances in Neural Information Processing Systems}.

\bibitem[{Buchanan and Feigenbaum(1978)}]{buchanan1978dendral}
Bruce~G Buchanan and Edward~A Feigenbaum. 1978.
\newblock Dendral and meta-dendral: Their applications dimension.
\newblock \emph{Artificial Intelligence}, 11(1-2):5--24.

\bibitem[{Chomsky(1957)}]{chomsky1957syntactic}
Noam Chomsky. 1957.
\newblock \emph{Syntactic Structures}.
\newblock Mouton de Gruyter.

\bibitem[{Donatelli et~al.(2021)Donatelli, Schmidt, Biswas, K{\"o}hn, Zhai, and Koller}]{donatelli2021aligning}
Lucia Donatelli, Theresa Schmidt, Debanjali Biswas, Arne K{\"o}hn, Fangzhou Zhai, and Alexander Koller. 2021.
\newblock Aligning actions across recipe graphs.
\newblock In \emph{Annual Conference on Empirical Methods in Natural Language Processing}.

\bibitem[{Feigenbaum(1981)}]{feigenbaum1981expert}
Edward~A Feigenbaum. 1981.
\newblock Expert systems in the 1980s.
\newblock \emph{State of the art report on machine intelligence. Maidenhead: Pergamon-Infotech}, 23.

\bibitem[{Fowler(2010)}]{fowler2010domain}
Martin Fowler. 2010.
\newblock \emph{Domain-specific languages}.
\newblock Pearson Education.

\bibitem[{Gao et~al.(2023)Gao, Madaan, Zhou, Alon, Liu, Yang, Callan, and Neubig}]{gao2023pal}
Luyu Gao, Aman Madaan, Shuyan Zhou, Uri Alon, Pengfei Liu, Yiming Yang, Jamie Callan, and Graham Neubig. 2023.
\newblock Pal: Program-aided language models.
\newblock In \emph{International Conference on Machine Learning}.

\bibitem[{Guizzardi(2013)}]{guizzardi2013ontology}
Giancarlo Guizzardi. 2013.
\newblock Ontology-based evaluation and design of visual conceptual modeling languages.
\newblock \emph{Domain Engineering: Product Lines, Languages, and Conceptual Models}, pages 317--347.

\bibitem[{Hopcroft et~al.(1996)Hopcroft, Motwani, and Ullman}]{hopcroft1996introduction}
John~E. Hopcroft, Rajeev Motwani, and Jeffrey~D. Ullman. 1996.
\newblock \emph{Introduction to Automata Theory, Languages, and Computation}.
\newblock Addison-Wesley Longman Publishing Co., Inc.

\bibitem[{Huang et~al.(2021)Huang, Li, Chen, Samel, Naik, Song, and Si}]{huang2021scallop}
Jiani Huang, Ziyang Li, Binghong Chen, Karan Samel, Mayur Naik, Le~Song, and Xujie Si. 2021.
\newblock Scallop: From probabilistic deductive databases to scalable differentiable reasoning.
\newblock In \emph{Advances in Neural Information Processing Systems}.

\bibitem[{Huang et~al.(2022)Huang, Abbeel, Pathak, and Mordatch}]{huang2022language}
Wenlong Huang, Pieter Abbeel, Deepak Pathak, and Igor Mordatch. 2022.
\newblock Language models as zero-shot planners: Extracting actionable knowledge for embodied agents.
\newblock In \emph{International Conference on Machine Learning}.

\bibitem[{Karsai et~al.(2009)Karsai, Krahn, Pinkernell, Rumpe, Schindler, and V{\"o}lkel}]{karsai2009design}
Gabor Karsai, Holger Krahn, Claas Pinkernell, Bernhard Rumpe, Martin Schindler, and Steven V{\"o}lkel. 2009.
\newblock Design guidelines for domain specific languages.
\newblock In \emph{OOPSLA Workshop on Domain-Specific Modeling (DSM' 09)}.

\bibitem[{Kiddon et~al.(2015)Kiddon, Ponnuraj, Zettlemoyer, and Choi}]{kiddon2015mise}
Chlo{\'e} Kiddon, Ganesa~Thandavam Ponnuraj, Luke Zettlemoyer, and Yejin Choi. 2015.
\newblock Mise en place: Unsupervised interpretation of instructional recipes.
\newblock In \emph{Annual Conference on Empirical Methods in Natural Language Processing}.

\bibitem[{Koupaee and Wang(2018)}]{koupaee2018wikihow}
Mahnaz Koupaee and William~Yang Wang. 2018.
\newblock Wikihow: A large scale text summarization dataset.
\newblock \emph{arXiv preprint arXiv:1810.09305}.

\bibitem[{Mernik et~al.(2005)Mernik, Heering, and Sloane}]{mernik2005and}
Marjan Mernik, Jan Heering, and Anthony~M Sloane. 2005.
\newblock When and how to develop domain-specific languages.
\newblock \emph{ACM Computing Surveys (CSUR)}, 37(4):316--344.

\bibitem[{Mikolov et~al.(2013)Mikolov, Sutskever, Chen, Corrado, and Dean}]{mikolov2013distributed}
Tomas Mikolov, Ilya Sutskever, Kai Chen, Greg~S Corrado, and Jeff Dean. 2013.
\newblock Distributed representations of words and phrases and their compositionality.
\newblock In \emph{Advances in Neural Information Processing Systems}.

\bibitem[{Momouchi(1980)}]{momouchi1980control}
Yoshio Momouchi. 1980.
\newblock Control structures for actions in procedural texts and pt-chart.
\newblock In \emph{COLING 1980 Volume 1: The 8th International Conference on Computational Linguistics}.

\bibitem[{Munaf{\`o} et~al.(2017)Munaf{\`o}, Nosek, Bishop, Button, Chambers, Percie~du Sert, Simonsohn, Wagenmakers, Ware, and Ioannidis}]{munafo2017manifesto}
Marcus~R Munaf{\`o}, Brian~A Nosek, Dorothy~VM Bishop, Katherine~S Button, Christopher~D Chambers, Nathalie Percie~du Sert, Uri Simonsohn, Eric-Jan Wagenmakers, Jennifer~J Ware, and John Ioannidis. 2017.
\newblock A manifesto for reproducible science.
\newblock \emph{Nature Human Behaviour}, 1(1):1--9.

\bibitem[{Papineni et~al.(2002)Papineni, Roukos, Ward, and Zhu}]{papineni2002bleu}
Kishore Papineni, Salim Roukos, Todd Ward, and Wei-Jing Zhu. 2002.
\newblock Bleu: a method for automatic evaluation of machine translation.
\newblock In \emph{Proceedings of the 40th annual meeting of the Association for Computational Linguistics}.

\bibitem[{Pareti et~al.(2014{\natexlab{a}})Pareti, Klein, and Barker}]{pareti2014semantic}
Paolo Pareti, Ewan Klein, and Adam Barker. 2014{\natexlab{a}}.
\newblock A semantic web of know-how: linked data for community-centric tasks.
\newblock In \emph{Proceedings of the International Conference on World Wide Web}.

\bibitem[{Pareti et~al.(2014{\natexlab{b}})Pareti, Testu, Ichise, Klein, and Barker}]{pareti2014integrating}
Paolo Pareti, Benoit Testu, Ryutaro Ichise, Ewan Klein, and Adam Barker. 2014{\natexlab{b}}.
\newblock Integrating know-how into the linked data cloud.
\newblock In \emph{Knowledge Engineering and Knowledge Management: 19th International Conference}.

\bibitem[{Puig et~al.(2018)Puig, Ra, Boben, Li, Wang, Fidler, and Torralba}]{puig2018virtualhome}
Xavier Puig, Kevin Ra, Marko Boben, Jiaman Li, Tingwu Wang, Sanja Fidler, and Antonio Torralba. 2018.
\newblock Virtualhome: Simulating household activities via programs.
\newblock In \emph{Conference on Computer Vision and Pattern Recognition}.

\bibitem[{Raffel et~al.(2020)Raffel, Shazeer, Roberts, Lee, Narang, Matena, Zhou, Li, and Liu}]{raffel2020exploring}
Colin Raffel, Noam Shazeer, Adam Roberts, Katherine Lee, Sharan Narang, Michael Matena, Yanqi Zhou, Wei Li, and Peter~J Liu. 2020.
\newblock Exploring the limits of transfer learning with a unified text-to-text transformer.
\newblock \emph{The Journal of Machine Learning Research}, 21(1):5485--5551.

\bibitem[{Russell(1923)}]{russell1923vagueness}
Bertrand Russell. 1923.
\newblock Vagueness.
\newblock \emph{The Australasian Journal of Psychology and Philosophy}, 1(2):84--92.

\bibitem[{Russell and Norvig(2010)}]{russell2010artificial}
Stuart~J Russell and Peter Norvig. 2010.
\newblock \emph{Artificial intelligence a modern approach}.
\newblock Prentice Hall Press.

\bibitem[{Schmid(1999)}]{schmid1999improvements}
Helmut Schmid. 1999.
\newblock Improvements in part-of-speech tagging with an application to german.
\newblock In \emph{Natural Language Processing using very large Corpora}.

\bibitem[{Schuff et~al.(2023)Schuff, Vanderlyn, Adel, and Vu}]{schuff2023human}
Hendrik Schuff, Lindsey Vanderlyn, Heike Adel, and Ngoc~Thang Vu. 2023.
\newblock How to do human evaluation: A brief introduction to user studies in nlp.
\newblock \emph{Natural Language Engineering}, pages 1--24.

\bibitem[{Shi et~al.(2023)Shi, Li, Niu, Xu, Liu, Xu, Gu, He, Li, Zhao et~al.}]{shi2023perslearn}
Yu-Zhe Shi, Shiqian Li, Xinyi Niu, Qiao Xu, Jiawen Liu, Yifan Xu, Shiyu Gu, Bingru He, Xinyang Li, Xinyu Zhao, et~al. 2023.
\newblock Pers{LEARN}: Research training through the lens of perspective cultivation.
\newblock In \emph{Annual Meeting of the Association for Computational Linguistics}.

\bibitem[{Tandon et~al.(2020)Tandon, Sakaguchi, Dalvi, Rajagopal, Clark, Guerquin, Richardson, and Hovy}]{tandon2020dataset}
Niket Tandon, Keisuke Sakaguchi, Bhavana Dalvi, Dheeraj Rajagopal, Peter Clark, Michal Guerquin, Kyle Richardson, and Eduard Hovy. 2020.
\newblock A dataset for tracking entities in open domain procedural text.
\newblock In \emph{Annual Conference on Empirical Methods in Natural Language Processing}.

\bibitem[{Tarski(1946)}]{tarski1946introduction}
Alfred Tarski. 1946.
\newblock \emph{Introduction to Logic and to the Methodology of Deductive Sciences}.
\newblock Dover Publications.

\bibitem[{Touvron et~al.(2023)Touvron, Lavril, Izacard, Martinet, Lachaux, Lacroix, Rozi{\`e}re, Goyal, Hambro, Azhar et~al.}]{touvron2023llama}
Hugo Touvron, Thibaut Lavril, Gautier Izacard, Xavier Martinet, Marie-Anne Lachaux, Timoth{\'e}e Lacroix, Baptiste Rozi{\`e}re, Naman Goyal, Eric Hambro, Faisal Azhar, et~al. 2023.
\newblock Llama: Open and efficient foundation language models.
\newblock \emph{arXiv preprint arXiv:2302.13971}.

\bibitem[{Wang et~al.(2023)Wang, Xu, Lan, Hu, Lan, Lee, and Lim}]{wang2023plan}
Lei Wang, Wanyu Xu, Yihuai Lan, Zhiqiang Hu, Yunshi Lan, Roy Ka-Wei Lee, and Ee-Peng Lim. 2023.
\newblock Plan-and-solve prompting: Improving zero-shot chain-of-thought reasoning by large language models.
\newblock \emph{arXiv preprint arXiv:2305.04091}.

\bibitem[{Wei et~al.(2022)Wei, Wang, Schuurmans, Bosma, Xia, Chi, Le, Zhou et~al.}]{wei2022chain}
Jason Wei, Xuezhi Wang, Dale Schuurmans, Maarten Bosma, Fei Xia, Ed~Chi, Quoc~V Le, Denny Zhou, et~al. 2022.
\newblock Chain-of-thought prompting elicits reasoning in large language models.
\newblock \emph{Advances in Neural Information Processing Systems}.

\bibitem[{Zhang et~al.(2023)Zhang, Xu, Yang, Zhou, You, Arora, and Callison-Burch}]{zhang2023causal}
Li~Zhang, Hainiu Xu, Yue Yang, Shuyan Zhou, Weiqiu You, Manni Arora, and Chris Callison-Burch. 2023.
\newblock Causal reasoning of entities and events in procedural texts.
\newblock In \emph{Findings of the Association for Computational Linguistics: EACL 2023}.

\bibitem[{Zhang et~al.(2012)Zhang, Webster, Uren, Varga, and Ciravegna}]{zhang2012automatically}
Ziqi Zhang, Philip Webster, Victoria~S Uren, Andrea Varga, and Fabio Ciravegna. 2012.
\newblock Automatically extracting procedural knowledge from instructional texts using natural language processing.
\newblock In \emph{LREC}.

\bibitem[{Zheng et~al.(2023)Zheng, Chiang, Sheng, Zhuang, Wu, Zhuang, Lin, Li, Li, Xing et~al.}]{zheng2023judging}
Lianmin Zheng, Wei-Lin Chiang, Ying Sheng, Siyuan Zhuang, Zhanghao Wu, Yonghao Zhuang, Zi~Lin, Zhuohan Li, Dacheng Li, Eric Xing, et~al. 2023.
\newblock Judging llm-as-a-judge with mt-bench and chatbot arena.
\newblock \emph{arXiv preprint arXiv:2306.05685}.

\bibitem[{Zhou et~al.(2022{\natexlab{a}})Zhou, Yin, and Neubig}]{zhou2022hierarchical}
Shuyan Zhou, Pengcheng Yin, and Graham Neubig. 2022{\natexlab{a}}.
\newblock Hierarchical control of situated agents through natural language.
\newblock In \emph{Proceedings of the Workshop on Structured and Unstructured Knowledge Integration}.

\bibitem[{Zhou et~al.(2022{\natexlab{b}})Zhou, Zhang, Yang, Lyu, Yin, Callison-Burch, and Neubig}]{zhou2022show}
Shuyan Zhou, Li~Zhang, Yue Yang, Qing Lyu, Pengcheng Yin, Chris Callison-Burch, and Graham Neubig. 2022{\natexlab{b}}.
\newblock Show me more details: Discovering hierarchies of procedures from semi-structured web data.
\newblock In \emph{Annual Meeting of the Association for Computational Linguistics}.

\end{thebibliography}

\clearpage
\onecolumn
\appendix
\renewcommand\thefigure{A\arabic{figure}}
\setcounter{figure}{0}
\renewcommand\thetable{A\arabic{table}}
\setcounter{table}{0}
\renewcommand\theequation{A\arabic{equation}}
\setcounter{equation}{0}
\pagenumbering{arabic}
\renewcommand*{\thepage}{A\arabic{page}}
\setcounter{footnote}{0}

\section{Additional remarks}

\subsection{\ad and expert systems}

The expert system is a highly structured representation of knowledge, and has been intensively studied while widely applied in the past few decades. Compared with the statistical model, however, it often requires more hand-crafted effort for development, and the workload increases dramatically with the increasing generality level. This echoes the primary goal of our proposed \ad framework, which makes efforts to formulate the structure of knowledge representation, similar to classic expert systems such as \citet{buchanan1978dendral,feigenbaum1981expert}, while adopting the power of data and \ac{llm} for domain generality and adaptation. 

\subsection{\ad framework and \ac{llm}}

The current \ad framework is essentially an \ac{llm}-in-the-loop approach. In the pre-processing of raw protocols, we take full advantage of \ac{llm} for conducting state-of-the-art \ac{ie}, obtaining key information from raw text, such as entities of operations, entities of conditions, and relations between different operations, \etc. Such key information is taken as the input of our core algorithms for syntactic constraint optimization and semantic constraint abstraction. 

We would like to reiterate that \ad is \textbf{not an alternative} for \ac{llm}. The proposed \ad framework is in essence yet another workflow for utilizing \ac{llm} to handle protocols with constraints. Compared with pure end-to-end workflow of \ac{llm} for producing such constraints, our framework is designed to be an explicit two-stage workflow. In the first stage, the framework automatically designs \acp{dsl} for target domains in a bidirectional optimization fashion. In the second stage, the resulting \acp{dsl} serve as plug-and-play constraint modules for all kinds of \acp{llm} to handle protocols precisely, preventing them from generating free-formed and non-determinstic procedural interpretations.

\section{On the generality of \ad}

\subsection{Generalize in the context of experimental protocols}

Regarding domains related to experimental sciences, there are syntactic constraints shared by domains and those unique to domains respectively, as we have discussed in \cref{subsec:algo-discuss}. These shared syntactic constraints mainly reflect the procedural nature across domains, and unique ones attributed to the composition types and the interruption in procedure execution. Through the following demonstrations, we aim to show that while semantic constraints are more diverse than syntactic constraints, there are still substantial distinctions between syntactic constraints across different domains.

\paragraph{Shared syntactic constraints across domain}

There are some examples of syntactic constraints that are shared across different areas: 
\begin{itemize}
    \item \texttt{allocate-statement}: \emph{Inoculate \textbf{5 ml LB medium} containing 100 µg/ml ampicillin with bacteria.} 
    \item \texttt{if-branch}: \emph{\textbf{If} cells are to be treated with PTX, divide them into two groups: one receives 100 ng/ml PTX.} 
    \item \texttt{parallel-for}: \emph{Iterate different MgCl2 concentrations \textbf{(1.5mM, 2.0mM, 2.5mM, and 3.0mM)} to find the optimal concentration for DNA amplification.} 
    \item \texttt{temporal-type-declaration}: \emph{Gently shake the reagent for \textbf{5 seconds} to mix.}
\end{itemize}

\paragraph{Unique syntactic constraints of one particular domain}

There are some examples of syntactic constraints that are unique to particular domains respectively:
\begin{itemize}
    \item In BioEng, \texttt{container-type-declaration}: \emph{Resuspend the pellet in PBS to refill the \textbf{tube}.}
    \item In Ecology, \texttt{string-type-declaration}: \emph{Close the soundproof chamber as instruction manual (see \textbf{"https://---"}).}
    \item In InfoBio, \texttt{raise-statement}: \emph{\textbf{If unbalanced spectral peaks}, noisy data, and frame loss \textbf{occur} during recording, ensure to subtract background noise and adjust for hemodynamic changes.}
\end{itemize}

\paragraph{Unique semantic constraints of one particular domain}

There are some examples of semantic constraints that are unique to particular domains respectively:

\begin{itemize}
    \item In Genetics, \begin{itemize}
        \item \texttt{DILUTE}: Dilute the Spike-inAmpR fragment to a concentration of 65.6 nM.
        \item \texttt{ATTACH}: Attach a Slide-A-Lyzer Float Buoy to the top (single) dialysis clip.
        \item \texttt{DELETE}: Detect the flow-through samples from each step with SDS-PAGE gel.
    \end{itemize}
    \item In Medical, \begin{itemize}
        \item \texttt{ASPIRATE}: Aspirate the DPBS and add 40 $\mu$03bcL of Sigma Lysis Solution for Blood.
        \item \texttt{ANESTHETIZE}: Anesthetize the rat intraperitoneally (i.p.).
        \item \texttt{DISSECT}: Dissect the fly brain under a stereomicroscope with light sources.
    \end{itemize}
    \item In Ecology, \begin{itemize}
        \item \texttt{STERILIZE}: Sterilize the seeds with 5\% (v/v) sodium hypochlorite.
        \item \texttt{QUANTIFY}: Quantify the eggs by placing washed eggs in a 10-mL graduated cylinder.
        \item \texttt{CALIBRATE}: Calibrate the motion sensor using a two-axis actuator.
    \end{itemize}
\end{itemize}

\subsection{Generalize beyond experimental protocols}

\ad focuses on a specialized form of natural language: procedural text, and is designed to formalize this procedural text, offering a clear execution trace and a readily verifiable interface. We select experimental sciences experiments for their demands on high interpretability, stringent execution, and adaptable planning. This methodology can be easily applied to new domains beyond experimental sciences, where such rigorous conditions are not necessary, including culinary recipes. Examples of various syntactic constraints are provided in \cref{tab:apdx-syncon-recipes}.

\begin{table}[h!]
    \centering
    \caption{\textbf{Showcases of syntactic constraints in culinary recipes}}
    \begin{tabular}{c|c}
    \toprule
    Constraint     &  Original recipe text  \\
    \midrule
    \texttt{integer-type-declaration}     &  8 [ounces] white fresh \{pasta\}   \\
    \texttt{device-type-declaration}     &  Yield: 2 plates \\
    \texttt{floatingpoint-type-declaration}  &  1/3 [cup] red \{wine\} \\
    \texttt{temporal-type-declaration}  & After |2 minutes| more, add the \@beef\@. \\
    \bottomrule
    \end{tabular}
    \label{tab:apdx-syncon-recipes}
\end{table}

We can also effectively design a \ac{dsl} with such a corpus where actions happen in a kitchen instead of a lab. We showcase a recipe example adopted from the Corel\footnote{Visit \url{https://fse.studenttheses.ub.rug.nl/25731/} for documentation.} \ac{dsl} as follows.

\begin{lstlisting}[]
Pasta Bolognese

Yield: 2 plates

Ingredients:

- 8 [ounces] white fresh {pasta}
- 1 [floz] olive {oil}
- 1/4 [ounce] {garlic}; minced
- 4 [ounces] {onions}; chopped
- 4 [ounces] shallow fried {beef}; minced
- 1 - 1 1/2 [ounce] lean prepared {bacon}
- 1/3 [cup] red {wine}
- 150 [gram] raw {carrots}; thinly sliced
- 2/3 [ounce] concentrated {tomato puree}
- 4 [ounces] red {sweet pepper}; cut julienne
- 1 [ounce] {parmesan} cheese

Instructions:

Add the @oil@ to a large saucepan, heat to <300 F>, and saute the @onions@. 

After |2 minutes|, add the @garlic@. 

Keep on medium to high heat, and don't stir. 

After |2 minutes| more, add the @beef@.

Fry the @bacon@ in a separate pan, on high heat. 

Remove liquified fat when done.

Boil @pasta@ in a medium pan, until al dente (~|8 minutes|). 

Drain when done.

Once the @beef@ is done, add the @carrots@, @sweet pepper@ and @tomato puree@. 

Slowly add the @wine@ as well, to not lower the temperature. 

Let it simmer (but not boil) for |5-10 minutes|.

Add the @bacon@ to the large saucepan.

Serve with grated @parmesan@ cheese.
\end{lstlisting}

Based on the example, it is trivial to figure out that the constraints on cooking recipes are essentially in a subset of those scientific procedures. The cooking procedures mainly focus on sequentially executing the actions without switching their execution orders. Also, most of the culinary ingredients are processed in single-thread fashion without duplication for different experimental groups. Hence, the Corel \ac{dsl} does not require some syntactic constraints for non-linear control flows, such as \texttt{loop}, \texttt{branch}, and \texttt{split}, while these constraints are dominant in \acp{dsl} for protocols. \ad will adaptively exclude these unnecessary features when optimizing the \ac{dsl} for culinary procedures given the corpus of recipes. Regarding semantics constraints, there are fewer shared semantics on operations, ingredients, and conditions between cooking and experimental sciences compared to those semantics shared between different domains within experimental sciences. However, this increment of semantic diversity would not yield significant challenge for \ad, as our semantic reduction does not rely on domain transfer.

\subsection{Extended discussions}
When considering both syntactic and semantic constraints across different domains, we observe that there are significant differences within the context of experimental sciences. However, when looking at a more general context that encompasses all possible procedural knowledge worldwide, these differences tend to converge. This suggests that \acp{dsl} may have a \textbf{hierarchical} structure in the general context. The shared features of different \acp{dsl} are more common closer to the root, such as the operation \texttt{incubate} shared by Genetics and Ecology in experimental sciences, and the operation \texttt{add} shared by Genetics, Ecology, and cooking. These common \ac{dsl} constructs possess general semantics. On the other hand, as we move closer to the leaf, the majority of \ac{dsl} constructs become specialized for unique domains. This parallels the spectrum of \emph{naturalism} in sciences~\citep{shi2023perslearn}, ranging from a general community with convergent high-level concepts to a specific community with divergent low-level knowledge.

\section{Implementation details}\label{sec:implement-detail}

\subsection{Implementation of the syntax prior model}\label{sec:Apdx-z-given-s}

We employ the \ac{cfg} generally for designing modern imperative programming language as the prior model $p(\mathcal{Z}|\mathcal{S})$ for syntax optimization.

\begin{lstlisting}[language=C]
<program> ::= <statements>


<statements> ::= <statement> 
                | <statement> <statements>

/* All support */ 
<statements> ::= <imperative-model>
               | <runtime-error-handling>
               | <type-system>
               | <concurrent>
               | <react>


/* Imperative Model */
<imperative-model> ::= <if-statement>
                      | <loop-statement>
                      | <jump-statement>
                      | <memory-management>
                      | <function-procedure>
                      | <arithmetic-expression>
                      | <logical-expression>
                      | <assignment-expression>

<if-statement> ::= "if" "(" <expression> ")" "{" <statements> "}" 
                  | "if" "(" <expression> ")" "{" <statements> "}" "else" "{" <statements> "}"

<loop-statement> ::= "While" "(" <expression> ")" "{" <statements> "}"
                    | "For" "(" <assignment-expression> ";" <expression> ";" <assignment-expression> ")" "{" <statements> "}"
                
<jump-statement> ::= "break"
                    | "continue" 
                    
<function-procedure>  ::= "Call" <identifier> "(" <arguments> ")"
                         | "Function" <identifier> "(" <parameters> ")" "{" <statements> "}"

<memory-management> ::= "allocate" <type> <identifier>
                       | "deallocate" <identifier>

<assignment-expression> ::= <identifier> "=" <expression>

<arithmetic-expression> ::= <expression> <arithmetic-operator> <expression>

<logical-expression> ::= <expression> <logical-operator> <expression>

/* Runtime Error Handling */
<runtime-error-handling> ::= <raise-stmt>
                            | <resolve-stmt>

<raise-stmt> ::= "raise" "(" <expression> ")"

<resolve-stmt> ::= "try" "{" <statements> "}" "catch" "(" <identifier> ")" "{" <statements> "}"

/* Type System */
<type-system> ::= <data-type>
                 | <class-type>
                 | <domain-specified-type>

<domain-specified-type> ::= "time"
                            | "reagent"
                            | "device"
                            | "container"
                            | <scientific-type>

<data-type> ::= "int"
               | "float"
               | "bool"
               | "string"
               | "set" "<" <type> ">"
               | "dict" "<" <type> "," <type> ">"
               | "vector" "<" <type> ">"

<class-type> ::= "class" <identifier> "{" <class-body> "}"

<kind-type> ::= "type" <identifier> "=" <type>

<class-body> ::= <class-members>

<class-members> ::= <class-member> | <class-member> <class-members>

<class-member> ::= <variable-declaration>
                  | <method-declaration>

<variable-declaration> ::= <assignment-expression>

<method-declaration> ::= <function-procedure>

/* Concurrent Programming */
<concurrent> ::= <data-parallel>
                | <message-passing>

<data-parallel> ::= "parallelFor" "(" <parallel-range> ")" "{" <statements> "}"
                   | "parallelMap" "(" <collection> "," <function> ")"

<parallel-range> ::= "range" "(" <expression> "," <expression> ")" // Define start, end of range

<collection> ::= <identifier> // Reference to a collection of data, e.g., array, list

<function> ::= <identifier> // Reference to a function to apply in parallel

<message-passing> ::= "spawnProcess" "(" <process-function> ")"
                     | "sendMessage" "(" <process-identifier> "," <message> ")"
                     | "receiveMessage" "(" <message-type> ")"

<process-function> ::= <identifier>

<process-identifier> ::= <identifier>

<message> ::= <expression>

<message-type> ::= <type>


/* React Model */
<react> ::= <event-stmt>
           | <response-stmt>

<event-stmt> ::= "emit" "(" <event> ")"

<response-stmt> ::= "on" "(" <event> ")" "{" <statements> "}"

/* Auxiliary Definitions */
<type> ::= <data-type> | <class-type> | <kind-type> | "void"

<parameters> ::= <empty> | <parameter> | <parameter> "," <parameters>

<parameter> ::= <identifier> ":" <type>

<arguments> ::= <empty> | <expression> | <expression> "," <arguments>

<arithmetic-operator> ::= "+" | "-" | "*" | "/"

<logical-operator> ::= "&&" | "||" | "!"

<identifier> ::= <letter> (<letter> | <digit>)*

<letter> ::= "A" | "B" | ... | "Z" | "a" | "b" | ... | "z"

<digit> ::= "0" | "1" | "2" | ... | "9"

<event> ::= <identifier>
\end{lstlisting}

\newpage
\subsection{Implementation of the prior belief function}\label{sec:Apdx-theta-given-s}

$p(\theta_k | \mathbf{c}_i)$ serves as a prior and represents the probability that programming language features $\theta_k$ are present within the experimental protocol $\mathbf{c}_i$.

\[
\resizebox{\textwidth}{!}{$p(\theta_k| \mathbf{c}_i) =
\begin{cases}
    \mathbf{max}_{i=1}^{L(\texttt{opc})} \text{Exists}(\texttt{INT}_i) & \text{if } \theta_k = \text{interger-type-declaration} \\

    \mathbf{max}_{i=1}^{L(\texttt{opc})} \text{Exists}(\texttt{FLOAT}_i) & \text{if } \theta_k = \text{floatingpoint-type-declaration} \\

    \mathbf{max}_{i=1}^{L(\texttt{opc})} \mathbf{max}(\text{Exists}(\texttt{TRUE}_i), \text{Exists}(\texttt{FALSE}_i)) & \text{if } \theta_k = \text{boolean-type-declaration} \\

    \mathbf{max}_{i=1}^{L(\texttt{opc})} \text{Exists}(\texttt{STRING}_i) & \text{if } \theta_k = \text{string-type-declaration} \\

    \mathbf{max}_{i=1}^{L(\texttt{opc})} \mathbf{max}\{\text{S}(\texttt{NOUN}_i, \texttt{"vector"})\} & \text{if } \theta_k = \text{vector-type-declaration} \\

    \mathbf{max}_{i=1}^{L(\texttt{opc})} \mathbf{max}\{\text{S}(\texttt{NOUN}_i, \texttt{"dictionary"})\} & \text{if } \theta_k = \text{dict-type-declaration} \\

    \mathbf{max}_{i=1}^{L(\texttt{opc})} \mathbf{max}\{\text{S}(\texttt{NOUN}_i, \texttt{"set"})\} & \text{if } \theta_k = \text{set-type-declaration} \\

    \mathbf{max}_{i=1}^{L(\texttt{opc})} \text{Exists}(\texttt{TIME}_i) & \text{if } \theta_k = \text{temporal-type-declaration} \\

    \mathbf{max}_{i=1}^{L(\texttt{opc})} \mathbf{max}\{\text{BelongTo}(\texttt{NOUN}_i, \texttt{Chem})\} =  \mathbf{max}_{i=1}^{L(\texttt{opc})} \text{Exists}(\texttt{CHEM}_i) & \text{if } \theta_k = \text{reg-type-declaration} \\

    1 & \text{if } \theta_k = \text{device-type-declaration} \\

    \mathbf{max}_{i=1}^{L(\texttt{opc})} \mathbf{max}\{\text{BelongTo}(\texttt{NOUN}_i, \texttt{Container})\} = \mathbf{max}_{i=1}^{L(\texttt{opc})} \text{Exists}(\texttt{CONTAINER}_i) & \text{if } \theta_k = \text{container-type-declaration} \\

    1 & \text{if } \theta_k = \text{scientific-type-declaration} \\
    
    \mathbf{max}_{i=1}^{L(\texttt{opc})} \text{S} (\texttt{opc}_i, \texttt{"repeat"}) & \text{if } \theta_k = \text{for-loop} \\

    \mathbf{max}_{i=1}^{L(\texttt{opc})} \text{S} (\texttt{opc}_i, \texttt{"repeat"}) \times \text{!Exists}(\texttt{INT}_i) & \text{if } \theta_k = \text{while-loop} \\

    \mathbf{max}_{i=1}^{L(\texttt{opc})} \text{Exists}(\texttt{IF}_i) & \text{if } \theta_k = \text{if-branch} \\

    \mathbf{max}_{i=1}^{L(\texttt{opc})} \text{Exists}(\texttt{IF}_i) \times \text{Exists}(\texttt{ELSE}_i) & \text{if } \theta_k = \text{if-else-branch} \\

    \mathbf{max}_{i=1}^{L(\texttt{opc})} \text{S} (\texttt{opc}_i, \texttt{"call"}) & \text{if } \theta_k = \text{function-procedure-call} \\

    \mathbf{max}_{i=1}^{L(\texttt{opc})} \text{S} (\texttt{opc}_i, \texttt{"call"}) & \text{if } \theta_k = \text{function-procedure-declaration} \\

    Q(\text{while-loop}| \mathbf{c}) \times Q(\text{if-branch}| \mathbf{c}_i) & \text{if } \theta_k = \text{break-statement} \\

    0 & \text{if } \theta_k = \text{continue-statement} \\

    \mathbf{max}_{i=1}^{L(\texttt{opc})} \mathbf{max}\{\text{BelongTo}(\texttt{NOUN}_i, \texttt{Chem}), 
    \text{BelongTo}(\texttt{NOUN}_i, \texttt{Container})\} & \text{if } \theta_k = \text{allocate-statement} \\

    \mathbf{max}_{i=1}^{L(\texttt{opc})} \text{Exists}(\texttt{ADD}_i) & \text{if } \theta_k = \text{add-arithmetic-operator} \\

    \mathbf{max}_{i=1}^{L(\texttt{opc})} \text{Exists}(\texttt{MINUS}_i) & \text{if } \theta_k = \text{minus-arithmetic-operator} \\

    \mathbf{max}_{i=1}^{L(\texttt{opc})} \text{Exists}(\texttt{MULTIPLY}_i) & \text{if } \theta_k = \text{multi-arithmetic-operator} \\

    \mathbf{max}_{i=1}^{L(\texttt{opc})} \text{Exists}(\texttt{DEVIDE}_i) & \text{if } \theta_k = \text{devid-arithmetic-operator} \\

    \mathbf{max}_{i=1}^{L(\texttt{opc})} \text{Exists}(\texttt{AND}_i) & \text{if } \theta_k = \text{and-arithmetic-operator} \\

    \mathbf{max}_{i=1}^{L(\texttt{opc})} \text{Exists}(\texttt{OR}_i) & \text{if } \theta_k = \text{or-arithmetic-operator} \\

    \mathbf{max}_{i=1}^{L(\texttt{opc})} \text{Exists}(\texttt{NOT}_i) & \text{if } \theta_k = \text{not-arithmetic-operator} \\

    \mathbf{max}_{i=1}^{L(\texttt{opc})} \text{Exists}(\texttt{EQUAL}_i) & \text{if } \theta_k = \text{assignment-expression} \\

    \mathbf{max}_{i=1}^{L(\texttt{opc})} \mathbf{max}\{\text{S}(\texttt{NOUN}_i, \texttt{"error"})\} & \text{if } \theta_k = \text{raise-statement} \\

    \mathbf{max}_{i=1}^{L(\texttt{opc})} \text{S} (\texttt{opc}_i, \texttt{"resolve"}) \times \mathbf{max}\{\text{S}(\texttt{NOUN}_i, \texttt{"error"})\} & \text{if } \theta_k = \text{resolve-statement} \\

    \exists \  x \in \texttt{Chem}: \sum_{i=1}^{L(\texttt{opc})}\text{Exists}(x \in \texttt{NOUN}_i) \geq \epsilon  & \text{if } \theta_k = \text{class-type-declaration} \\

    \mathbf{max}_{i=1}^{L(\texttt{opc})} \text{Exists}(\texttt{PERSONS}_i)  & \text{if } \theta_k = \text{spawn-process} \\

    Q(\text{spawn-process}| \mathbf{c})  \times \mathbf{max}_{i=1}^{L(\texttt{opc})} \text{S} (\texttt{opc}_i, \texttt{"say"}) & \text{if } \theta_k = \text{send-message} \\

    Q(\text{spawn-process}| \mathbf{c})  \times \mathbf{max}_{i=1}^{L(\texttt{opc})} \text{S} (\texttt{opc}_i, \texttt{"say"})  & \text{if } \theta_k = \text{receive-message} \\

    \mathbf{max}_{i=1}^{L(\texttt{opc})} \text{Exists}(\texttt{WHEN}_i) & \text{if } \theta_k = \text{event-statement} \\

    \mathbf{max}_{i=1}^{L(\texttt{opc})} \text{Exists}(\texttt{WHEN}_i) & \text{if } \theta_k = \text{response-statement} \\

    \mathbf{max}_{i=1}^{L(\texttt{opc})} \text{Exists}(L(\texttt{NUM}_i) \geq \epsilon) & \text{if } \theta_k = \text{parallel-for} \\

    \mathbf{max}_{i=1}^{L(\texttt{opc})-1} \mathbf{max}_{j=i+1}^{L(\texttt{opc})} \ \text{Exists}(\texttt{CHEM}_i) \times \text{Exists}(\texttt{CHEM}_i=\texttt{CHEM}_j) \times \\ \text{!Exists}(\texttt{PRON}_i) \times \text{!Exists}(\texttt{PRON}_j) \times (1 - \text{S} (\texttt{Sentence}_i, \texttt{Sentence}_j))\\& \text{if } \theta_k = \text{parallel-map}
\end{cases}$}
\]

In the context of our prior belief function, $\texttt{Chem}$ and $\texttt{Container}$ are predefined sets. The set $\texttt{Sentence}$ includes all possible sentences, each of which begins with an opcode. These opcodes together constitute the set "$\texttt{opc}$." In our algorithm, we assign $\epsilon$ with 4 and assign $\delta$ with 0.5.

We derive these various sets from filtering based on specific rules.

For pre-processing, We utilize regular expressions to match a variety of elements, such as integers, floating-point numbers, boolean values, strings, and control flow elements. Then, we employ speech tagging to label nouns and pronouns~\citep{schmid1999improvements}. Afterwards, we conduct entity recognition for the identification of chemical elements. 

To determine word similarity, we employ word embedding model to calculate cosine similarities~\citep{mikolov2013distributed}. To assess sentence similarity, pairwise word similarity scores are averaged across the words in the sentences. Furthermore, we use GPT-3.5 to extract initial specific scientific data types. Subsequently, the preliminary extraction results are filtered through a list of candidate words to obtain the final results.

The prompt is as follows:

\begin{lstlisting}[]
You need to identify and list any physical chemistry scientific quantities mentioned in the following experimental procedure.
Requirements:
1. The output format should be only one per line as "Original Text: Scientific Quantity".
2. Please extract information from the given sentences without creating your own summaries.
Output Example: 
50mL: Volume
{}
The given experimental procedure are:
{}
\end{lstlisting}

The filtering word list is as follows:

\begin{lstlisting}[]
"Volume", "Temperature", "Length",
"Energy", "Concentration", "Mass", "Speed", "Acceleration", "Density", "Frequency",
"Force", "Acidity", "Flow Rate", "Pressure", "Voltage"
\end{lstlisting}

\subsection{Implementation of feature vector transformation}\label{sec:Apdx-feature-vector}

Each $\mathbf{x}_i$ in the dataset $\mathcal{X} = \{\mathbf{x}_1, \mathbf{x}_2, ..., \mathbf{x}_N\}$ is a feature vector representing an operation pattern extracted from the corpora. To obtain this dataset, we extract all sentences starting with verbs from the dataset using NLTK's part-of-speech tagging firstly. We then utilize GPT to annotate the parameters within these sentences and ultimately convert them into ont-hot feature vectors. 

The prompt is as follows:

\begin{lstlisting}[]
You need to complete an entity recognition task with defined entity categories: {}
Requirements:
1. The output format should be annotated on the original sentence, and only the annotated sentence should be output.
2. Please extract information from the given sentences without creating your own summaries.
3. Text is in a laboratory setting, please carefully analyze the specialized terms in the fields of biology and chemistry.
4. Please extract as many entities as possible from this sentence.
Example:
{}
The given sentences are:
{}
\end{lstlisting}

\subsection{Implementation of \ac{dpmm}}\label{sec:Apdx-dpmm}

\paragraph{Spectral clustering via \ac{dpmm}}

The \ac{dpmm} facilitates the spectral clustering of these feature vectors into groups representing unique atomic semantic constraints, accommodating the variability and complexity of semantic patterns across different domain corpora. The iterative clustering process, centered around Gibbs sampling, estimates the posterior distributions of the model's parameters. This not only identifies distinct semantic constraints but also adapts to the intricate semantic patterns prevalent in various domain-specified corpora.

\paragraph{Algorithm symbols and initialization}

Key symbols include $\mathcal{X}$ for the dataset of feature vectors; $N$, the total number of samples; and $L$, the flexible number of clusters within the mixture model. Latent variables $\mathcal{W}$ indicate cluster assignments for each $\mathbf{x}_i$, while $\Psi$ encompasses the set of parameters for each cluster. The concentration parameter $\alpha$ influences new cluster formation likelihood, with $\sigma_m$ serving as the regularization term for parameter updates.

Initially, $K=1$ and cluster parameters $\psi_0$ are set with generic values, \eg, $\mu=0$, $\sigma=1$, assigning all data points to an initial cluster ($w_i=0$ for each $i$).

\paragraph{Gibbs sampling and posterior inference}

The core of the algorithm lies in the Gibbs sampling loop, which iterates until convergence. It updates the cluster assignment $w_i$ for each $\mathbf{x}_i$ by evaluating the posterior distribution that incorporates the current parameter estimates and $\alpha$. The posterior probability of $\mathbf{x}_i$ belonging to cluster $l$ integrates the likelihood of observing $\mathbf{x}_i$ given the cluster parameters and the prior probability of cluster membership, formalized as:
\begin{equation}
p(w_i = l | \mathbf{w}_{-i}, \mathbf{x}_i, \alpha, \Psi) \;\propto\; p(\mathbf{x}_i | \psi_l) \cdot p(w_i = l | \mathbf{w}_{-i}, \alpha),
\end{equation}
where $p(\mathbf{x}_i | \psi_l)$ is the likelihood of observing $\mathbf{x}_i$ under the parameters of cluster $l$, and $p(w_i = l | \mathbf{w}_{-i}, \alpha)$ reflects the adjusted prior probability for existing clusters and the potential for new cluster formation, influenced by $\alpha$.

Following the assignment update, cluster parameters $\Psi$ are recalibrated using \ac{map}, applying regularization with $\sigma_m$ to ensure stability and prevent overfitting.

\subsection{Exemplar output}

There are several examples demonstrating the abstraction from natural language instructions to semantic patterns, from domains of Medical, Genetics, and Ecology.

Medical:
\begin{lstlisting}[]
{
    "TRANSECT": [
        {
            "pattern": ["REG", "REG"],
            "example": [
                "Transect the [aorta]{REG} proximally to the origin of the [brachiocephalic trunk]{REG}.",
                "Transect the [bile duct]{REG} close to the [pancreas]{REG}.",
                "Transect the [SHVC]{REG} along with part of the diaphragm and [IHVC]{REG} at the level of the left renal vein when the donor liver became pallid.",
                "Transect the liver parenchyma of the [caudate lobe]{REG} and the [Spiegel lobe]{REG}.",
                "Transect the [infrahepatic inferior vena cava (IHIVC)]{REG} and mobilize the [lVC]{REG} from retroperitoneal tissue."
            ],
        },
        {
            "pattern": ["REG", "Device"],
            "example": [
                "Transect the [cranial nerves]{REG} with the [scissors]{Device}.",
                "Transect the [PHA]{REG} of the recipient at its root to expose the vascular lumen using [micro-scissors]{Device}.",
                "Transect the [femoral artery]{REG} in the section located between the distal and proximal knots using [spring scissors]{Device}."
            ],
        },
        {
            "pattern": ["REG", "Device", "REG"],
            "example": [
                "Transect the [aorta]{REG} with [fine sharp scissors]{Device} (see Table of Materials) just proximal to the [subclavian artery]{REG} takeoff."
            ],
        },
        {
            "pattern": ["REG"],
            "example": [
                "Transect the [colon]{REG}.",
                "Transect the [umbilical ligament]{REG}.",
                "Transect the [IVC]{REG} 1 cm above the diaphragm.",
                "Transect the [stretcher/opener motor nerve]{REG}."
            ],
        }
    ]
}
\end{lstlisting}

Genetics:
\begin{lstlisting}[]
{
    "RESOLVE": [
        {
            "pattern": ["REG", "REG", "REG", "Length"],
            "example": [
                "Resolved 1 \u00b5l of the [PCR reaction]{REG} on a [DNA gel]{REG} to confirm the successful reaction with a [DNA band]{REG} at about [10 kb]{Length}.",
                "Resolve 2 \u00b5l of the [PCR reaction]{REG} on a [DNA gel]{REG} to confirm the successful reaction to see if a [PCR product]{REG} of [3.5 kb]{Length} is present."
            ],
        },
        {
            "pattern": ["Concentration", "Concentration"],
            "example": [
                "Resolve [0.05%]{Concentration} of the input (=extract/lysate) and at least [10%]{Concentration} of the eluates on a 4\u201315% Mini-PROTEAN\u00ae TGX\u2122 Precast SDS-PAA gel."
            ],
        },
        {
            "pattern": ["REG", "REG"],
            "example": [
                "Resolve the cell pellet in [PBS]{REG} and [centrifuge]{REG} again.",
                "Resolve [5 \u00b5L]{Volume} of the [PCR]{REG} from [6-12]{String} representative samples on a [2% (w/v)]{Concentration} [agarose gel]{REG}.",
                "Resolve [10 \u00b5l]{Volume} of [PCR reactions]{REG} on [1%]{Concentration} [agarose gel]{Container} ([TAE]{REG} or [TBE]{REG}) with a [1 kb]{Length} [DNA ladder]{REG} to check for positive amplification ([Figure 3C]{String})."
            ]
        }
    ]
}
\end{lstlisting}

Ecology:
\begin{lstlisting}[]
{
    "DRILL": [
        {
            "pattern": ["Device"],
            "example": [
                "Drill a hole big enough to allow the [sensor]{Device} to be inserted.",
                "Drill a hole at this coordinate using a [medium tip burr]{Device} (Figure 3)."
            ],
        },
        {
            "pattern": [
                "Length"
            ],
            "example": [
                "Drill a hole (approximately [7 mm]{Length}) in the skull above the target area using coordinates from a brain atlas according to the animal used.",
                "Drill a craniotomy of approximately [1 mm]{Length} in diameter in this location.",
                "Drill the [craniotomy]{Length}."
            ],
        },
        {
            "pattern": [
                "REG"
            ],
            "example": [
                "Drill out the hole in the [pulley]{REG} to match the diameter of the [motor shaft]{REG}.",
                "Drill the [skull]{REG} on the marked point using a .9 mm diameter steel drill tip."
            ],
        },
        {
            "pattern": [
                "Length",
                "Device"
            ],
            "example": [
                "Drill a [3 in]{Length} hole in the top panel for the [fan]{Device}.",
                "Drill a small burr hole (approximately [0.5 mm]{Length}) using the [Micro-Drill]{Device} at the [coordinates]{REG}."
            ],
        }
    ],
}
\end{lstlisting}

\subsection{Extended discussions}

\paragraph{On the assignments of $K$ and $K'$} 

Empirically, $K'$ and $K$ are in the same order of magnitude, with a difference roughly ranging in $25\% \sim 35\%$, in the experiments we have conducted. Intuitively, $K'$ denotes an estimation of syntactic constraints before optimization and $K$ is the exact total number of syntactic constraints after optimization.

$K'$ is specified by an appropriate initialization. It is grounded to a predefined upper bound according to the size of the general syntax tree for C/C++. During the optimization process, the syntax tree is pruned according to the observed distribution of syntactic constraints on the corresponding domain corpus.

\paragraph{On the scale of training corpus}

In our studies, we have found that approximately 348 protocols, with an average of 121 steps per protocol, totaling 42,108 steps, are needed as a minimum amount of domain-specific corpus for the \ad framework to work on a specific domain and to design a \ac{dsl} accordingly. This threshold ensures that the dataset possesses sufficient richness of domain-specific context and is robust for extracting syntax and semantic constraints.

\section{Data sources}

\subsection{Corpora}\label{sec:s-data-corpora}

Our corpora come from five websites including Nature\footnote{\url{https://protocolexchange.researchsquare.com/}}, Cell\footnote{\url{https://star-protocols.cell.com/}}, Bio\footnote{\url{https://bio-protocol.org/en}}, Wiley\footnote{\url{https://currentprotocols.onlinelibrary.wiley.com/}} and Jove\footnote{\url{https://www.jove.com/}}. From a domain perspective, our corpora are classified into 5 domains including BioEng, InfoBio, Medical, Ecology and Genetics. 
We quantified each domain corpus in Table~\ref{tab:corpus} by assessing the website sources, total number of protocols, and the average number of steps, actions, and objects per protocol. We only use sentences with subject-verb-object and predicate-object constructions to count the steps, actions, and objects. In addition, we performed the part-of-speech tagging of each word in the sentence, treating the verb as an action and the noun as an object. To note, we take the prototype of the verb or noun for the statistics.

\begin{table}[ht]
    \centering
    \caption{\textbf{Statistics of the corpora across five domains}}
    \begin{tabular}{cccccccccc}
        \toprule
         \textbf{Domain}     & \textbf{Nature}   & \textbf{Cell} & \textbf{Bio} & \textbf{Wiley} & \textbf{Jove} & \textbf{total} & \textbf{step} & \textbf{action type} & \textbf{object type}\\ 
        \midrule
        \textbf{InfoBio}     & 0  & 298  & 0 & 50 & 0 & 348 & 121 &  120 &  280  \\ 
        \textbf{Ecology}     & 55  & 166  & 0 & 34 & 557 & 812 & 71 &  88 &  193  \\ 
        \textbf{BioEng}     & 941  & 1404  & 0 & 22 & 1230 & 3597 & 70 &  80 &  188  \\
        \textbf{Medical}     & 290  & 1530  & 1061 & 116 & 4354 & 7351 & 65 &  80 &  189  \\ 
        \textbf{Genetics}     & 1045  & 2296  & 3522 & 134 & 1797 & 8794 & 58 &  73 &  175  \\ 
        \bottomrule
    \end{tabular}
    \label{tab:corpus}
\end{table}

We demonstrate procedure cases from five domains with respect to sources.

\begin{lstlisting}[]
[BioEng/Nature]
1. Virulent S. aureus MB 2865 \(Smith strain) was grown overnight with aeration in TSB medium.  
2. Bacteria were harvested and washed with fresh TSB and serially diluted to achieve an infective inoculum of 4 x 10<sup>3</sup> CFU/mL in 5% hog gastric mucin, and 0.5 mL was used to infect intraperitoneally.  
3. The infected mice were anesthetized using isoflurane \(Abbott).  0.5 inch hollow connectors were attached to the external section of the JVC catheter and the mice were subsequently harnessed to the counter-balance arm inside a mouse cage.  
4. Fluid primed PE 20 tubings attached to individual syringes on the infusion pump were connected to the mice through the 0.5 inch hollow connector.  
5. The infusion pump was initiated for 24 hours at a flow rate of 0.1 mL/hour.  
6. Blood samples are taken by tail vein to determine serum drug concentration during the infusion.  
7. After the 24 hour period, the mice were euthanized and the kidneys were aseptically removed and homogenized.  
8. Serial dilutions of the homogenates were plated on Mannitol plates and incubated overnight at 37 °C.  
9. Bacterial counts were enumerated \(ref. 2).
\end{lstlisting}

\begin{lstlisting}[]
[InfoBio/Cell]
Timing: 1 h per experiment (for step 3)
      The copulation-triggered photostimulation system detects copulation events
      in real time and turns on the green laser required for GtACR1-mediated
      inhibition of neural activity. This system allows for the inhibition of
      specific neural circuits in males only during copulation. In this assay,
      we use males that express GtACR111[href=https://www.wicell.org#bib10] in
      piezo-expressing neurons (w1118; +/+; piezo-GAL4/UAS-GtACR1.d.EYFP). As a mating partner, we use
      wild-type females. See
      key resources table[href=https://www.wicell.org#key-resources-table] for details on fly
      strains.
        Prepare GtACR1-expressing male flies.
            Set up the fly crosses. Introduce parental flies (4-5
            piezo-GAL4 males and UAS-GtACR1 virgin females) into
            vials containing fly media. The detailed genotypes are listed in the
            key resources table[href=https://www.wicell.org#key-resources-table].
          
Transfer parental flies (F0 fly) to fresh vials every 5 days.
            Collect F1 males of the desired genotypes according to step 1 in
            before you begin[href=https://www.wicell.org#before-you-begin].
          
Note: F0 flies are transferred to fresh
      vials up to 2 times.
    
Note: Set-up several vials at a time to
      obtain enough F1 male flies.
    
Note: Wild-type females are prepared as
      mating partners according to step 1 in Before you begin.
    
Pause point: The experimenter can
      decide the timing of subsequent feeding of all-trans-retinal (ATR) as long
      as the tested flies are within 5-8 days after eclosion at the time of the
      optogenetic assay.
        Feeding of ATR. Male flies of the experimental group are fed food
        containing ATR. Control males are fed food containing ethanol (EtOH)
        solvent.
Prepare plastic tubes containing ~150 μL of fly food.

            For the experimental group, add 2 μL of ATR, 25 mg/mL dissolved in
            99.5% EtOH (need to be prepared just before use) on the food
surface. For the control group, add 2 μL of 99.5% EtOH on the food
            surface instead of ATR.
            Transfer males to a plastic tube containing fly food with the
            surface covered with ATR (or EtOH) 24-36 h before the experiment.
            Keep the male flies on prepared food for 24-36 h in the dark before
            being used for the assays.
\end{lstlisting}

\begin{lstlisting}[]
[Medical/Bio]
Spleens were obtained from naive and malaria-infected C57BL/6 mice and were homogenized at room temperature in 6-well plates with 6 ml of HBSS with 2% FCS through a 70 μm cell strainer to form single-cell suspensions.   Red blood cells were lysed using 1 to 2 ml of RBC lysing buffer (depending on size of spleen) and splenocytes were washed once at 200 x g 4 °C for 5 min with cold FACS buffer (HBSS with 2% FCS).    Splenocytes were resuspended by gentle tapping on a rack in FACS buffer and kept on ice at all times to avoid background phosphorylation of STAT proteins.    Viability and cell counts were obtained by trypan blue exclusion using a haemocytometer.    Cells were washed once with 1 ml AIM V® Medium, resuspended at 20 x 106 cells/ml and rested on ice for a minimum of 20 min.    1 × 106 cells were incubated with 20 ng/ml rIL-2 or 2.5 ng/ml rIL-12 for 10 min at 37 °C, 5% CO2 (final volume of 200 μl) and immediately fixed on ice for 15 min by adding an equal volume of 4% paraformaldehyde.    Cells were washed with FACS buffer, resuspended in 500 μl of 90% ice-cold methanol and immediately stored down at -20 °C for a minimum of 2 h (cells can be kept for up to a month without affecting further staining).    Splenocytes were washed twice with FACS buffer and stained in FACS buffer at room temperature for 30 min for CD4 (GK1.5), CD44 (IM7), CD62L (MEL-14), T-bet (4B10) and phosphorylated STAT4 (at residue Y693, clone 38) or phosphorylated STAT5 (at residue Y694, clone 47).    Cells were washed with FACS buffer and analysed by flow cytometry.    Fluorescence minus one controls were included to validate flow cytometric results. Flow cytometry acquisition was performed using an LSR II.
\end{lstlisting}

\begin{lstlisting}[]
[Ecology/Wiley]
Generating primary transformants
1. Sow seeds of the transformation on GM medium containing gentamicin to select for transgenic plants harboring the T-DNA.
2. Let the seeds maturate for 2 weeks at 22°C in a growth chamber.
3. For GT: Pick at least 40 T1 plants containing pDe-EC-ttLbCas12a and transfer them to soil until maturity.
4. For cleavage activity: Extract DNA from 20 primary transformants containing the pDe-ttLbCas12a construct via the rapid DNA extraction method (see Edwards et al., 1991[href=https://currentprotocols.onlinelibrary.wiley.com/doi/10.1002/cppb.20117#cppb20117-bib-0002]). Analyze the mutagenesis efficiency by e.g., TIDE (see Critical Parameters).
5. Extract DNA from one leaf of each plant and set up a suitable PCR to verify the presence of your construct.
Obtaining heritable GT plants
6. Harvest the seeds of each T1 plant separately.
7. Sow about 100 seeds per line on GM medium.
8. Let the seeds maturate for 2 weeks at 22°C.
9. Extract DNA from one leaf of 100 plants per line as a pool.
10. Screen for positive GT events using a suitable PCR (see Critical Parameters).
11. Extract DNA from each plant of the positive identified T1 pools separately after another week of growth.
12. Analyze the T2 plants separately for heritable GT via PCR and confirm it by sequencing (see Critical Parameters).
13. Transfer positive T2 plants to soil and cultivate them to maturity.

\end{lstlisting}

\begin{lstlisting}[]
[Genetics/Jove]
All experiments involving the differentiation of human iPSC lines were performed in compliance with the Institutional Review Board of Boston University (protocol H33122). The dermal fibroblasts, procured for reprogramming to iPSCs, were obtained from a donor with written informed consent, under the approval of the Human Research Protection Office of Washington University School of Medicine, St. Louis, MO. Reprogrammed iPSCs were generated at the Center for Regenerative Medicine at Boston University and Boston Medical Center, Boston, MA.
1. Alveolosphere dissociation
Prepare complete serum-free differentiation media (cSFDM) as per the composition mentioned in Table 1.
Prepare CK + DCI media in the prepared cSFDM base as per Table 2.
Thaw 2D (human embryonic stem cell-qualified) and/or 3D (growth-factor reduced) matrix on ice as required for the experimental needs.
Aspirate all the CK + DCI medium using a pipette or aspirating pipette with vacuum from the 3D matrix droplets containing alveolospheres, derived from directed differentiation19, in a 12-well plate.
Add 1 mL of dispase (2 mg/mL) per droplet. Gently pipette the droplet into the dispase using a P1000 pipette. Incubate at 37 °C for 1 h, pipetting up and down once after 30 min.
Transfer the dissociated organoids (from Step 1.5) from one matrix droplet in the dispase to a 15 mL conical tube. To wash, add 10 mL of Iscove's Modified Dulbecco's Medium (IMDM, see Table of Materials).
Centrifuge at 300 x g for 5 min at room temperature. Aspirate the supernatant using a pipette or aspirating pipette with vacuum, leaving as little supernatant as possible.
NOTE: It is important to remove all dispase as any remaining dispase may dissolve the matrix that the cells will subsequently be seeded into. If a clear haze is seen above the pellet, the dispase has not completely dissolved the matrix, and more dispase can be added to the pellet for another 20-30 min at 37 °C.
Resuspend the cells in 1 mL of 0.05% trypsin per droplet and transfer back to the 12-well plate. Incubate at 37 °C for 12-15 min. Observe the dissociation under a microscope. Avoid over-pipetting the cells at this stage.
        NOTE: At the end of incubation, the cells need to achieve a single-cell suspension after pipetting 3-5 times with a P1000 pipette. For passaging iAT2s to ALI (Step 3), the trypsinization time needs to be minimized (maximum 12 min), such that the cells are in 2-3-cell clumps rather than single-cell suspension when ready for plating onto the cell culture insert.
Stop the action of trypsin with an equal volume of FBS-containing medium (10% ES-qualified FBS in DMEM). Centrifuge at 300 x g for 5 min at room temperature.
Wash the cells with 10 mL of IMDM. Centrifuge at 300 x g for 5 min at room temperature.
Resuspend the cells in an appropriate volume for counting, and then count the cells using a hemocytometer (see Table of Materials).
        NOTE: From one confluent 50 µL matrix droplet seeded at 400 cells/µL, the expected yield is 500,000 to 1.5 x 106 cells per droplet.
Use the single-cell suspension of iAT2 cells to generate alveolospheres by plating in the 3D matrix (Step 2) and/or plating on cell culture inserts for ALI culture (Step 3).
2. 3D plating of iAT2s
After counting (Step 1.11), determine the number of desired cells to replate in the 3D matrix (400 cells/µL of the matrix with 50-100 µL of 3D matrix droplets per well of a 12-well plate). Centrifuge the cells at 300 x g for 5 min at room temperature. Remove as much supernatant as possible using a pipette.
Resuspend the cells in the 3D matrix. Resuspend quickly and on ice, if needed, to prevent the matrix from polymerizing (which occurs when warm).
Use a P200 pipette to dispense one 3D matrix droplet per well into a pre-warmed 12-well plate. Pipette carefully to avoid creating bubbles in the matrix droplet. Do not allow the cell suspension to settle while dispensing multiple droplets.
Place the plate in a 37 °C incubator for 20-30 min to allow the matrix droplets to polymerize.
Add 1 mL of CK + DCI + 10 µM of Y-27632 medium (see Table of Materials) per well to cover the matrix droplet.
After 72 h, change the medium to CK + DCI without 10 µM of Y-27632.
Replace the medium with fresh CK + DCI every 48-72 h.
        NOTE: iAT2s will typically need to be passaged approximately every 10-14 days, depending on cell line and plating density.

\end{lstlisting}

\subsection{Data preprocessing}\label{sec:s-data-preproc}

We perform some preprocessing on the source data, including slice procedures and domain mapping.

\paragraph{Slice procedures} 

We segment the procedures of the protocols without destroying their structure as much as possible. Specifically, we first split the procedures according to the regular delimiter '[$\backslash$n$\backslash$r]+' to obtain a series of sub-paragraphs. Then we merge the sub-paragraphs in order, and the token number of merged paragraphs does not exceed the predetermined maximum token (i.e., 300). If the number of tokens in the merged paragraph is never satisfied, we follow the separator '.?!' to split and re-merge the paragraphs at a fine-grained level.

\paragraph{Domain mapping} 

The protocols from the five data sources have their own subject areas. We mapped the topic domains to the five domains developed by the experts according to the expert-set mapping table.
\begin{lstlisting}[]
[Genetics]
Antibody
Biochemistry
Biomarkers
Biophysics
Cell Biology
Cell Differentiation
Cell Membrane
Chromatin Immunoprecipitation (ChIP)
Gene Expression
Genetics
Genomics
Human Genetics
Molecular Biology
Molecular/Chemical Probes
Mouse Biology
Protein Biochemistry
Protein Science
Proteomics
RNaseq
Sequence Analysis
Sequencing
Signal Transduction
Single Cell
Single-Molecule Assays
Structural Biology
Systems Biology
Microbiology
Developmental Biology
Model Organisms
-----------------------------------------
[Medical]
Cancer
Cancer Research
Cardiology
Diseases
Drug Discovery
Gastroenterology
Health Sciences
Immunology
In Situ Hybridization
lmmunology and Infection
Medicine
Neuroscience
Oncology
Organoids
Physiology
Pharmacology
Psychology
Rheumatology
Stem Cells
Stem Cell Biology
Tissue Engineering
Toxicology
Urology
Health Humanities
-----------------------------------------
[Ecology]
Behavior
Ecology
Plant Sciences
Plant Biology
-----------------------------------------
[Bioeng]
Bioengineering
Biotechnology
Cell Culture
Microscopy
Biological Techniques
Cell Isolation
Cell Separation/Fractionation
Cell-Based Assays
Chemical Biology
Cytometry
Mass Cytometry
Protein Expression And Purification
-----------------------------------------
[Infobio]
Bioinformatics
Computational Biology and Bioinformatics
\end{lstlisting}

\section{Constraint design assessment details}

\subsection{Meta evaluation}

For meta evaluation, the experts' participation will include completing a survey and possibly partaking in an interview. The survey will present a series of statements or questions about the \ac{dsl}. For each item, we request two responses from the experts: Opinion on a 1-5 Scale and Confidence in the Response on a 1-5 Scale. This dual-scale approach will help quantify both the perspective on the \ac{dsl} and confidence in each response, providing a richer dataset for analysis.

In this part, we organized three experts in the field of experimental sciences to independently score our \ac{dsl}. Each expert scored the results of a single \ac{dsl} output across 9 dimensions, with a total of 50 questions. This process will be repeated 5 times, with experts independently evaluating 5 different subsets. We collected the scores from the three experts for subsequent processing and comparison.

The participants are instructed by the following guidance:
\begin{lstlisting}[]
Invitation to Participate in the Evaluation of a New Domain-Specific Language for Experimental Sciences

Dear Expert,

We are reaching out to invite you to partake in a crucial evaluation of a newly developed Domain-Specific Language (DSL) designed for life sciences. Your expertise in this field is invaluable for this assessment.

Purpose of the Experiment:
Our team has developed a DSL to facilitate both computerized and manual experimentation processes in life sciences. This evaluation aims to gather detailed feedback from experts like you to refine the DSL's usability and functionality.

What Will Be Involved:
Your participation will include completing a survey and possibly partaking in an interview. The survey will present a series of statements or questions about the DSL. For each item, we request two responses from you:
   1. Opinion on a 1-5 Scale: Rate each item on a scale where 1 indicates 'Strongly Disagree' and 5 indicates 'Strongly Agree'. The scale is nuanced as follows:
  - 1: Strong Disagreement or Major Issues
  - 2: Disagreement or Notable Concerns
  - 3: Neutral or Mixed Feelings
  - 4: Agreement or Minor Concerns
  - 5: Strong Agreement or Highly Favorable
   2. Confidence in Your Response on a 1-5 Scale: Indicate your level of confidence in your response to each item, where 1 is 'Not Confident at All' and 5 is 'Extremely Confident'. The scale implies:
  - 1: Very Low Confidence
  - 2: Low Confidence
  - 3: Moderate Confidence
  - 4: High Confidence
  - 5: Very High Confidence

This dual-scale approach will help quantify both your perspective on the DSL and your confidence in each response, providing a richer dataset for analysis.

In the interview, we will delve deeper into your experiences with the DSL, allowing you to share more comprehensive insights and suggestions.

Estimated Time Commitment:
The survey should take approximately 60 minutes to complete. The interview, if you choose to participate, will be a 1-hour session, scheduled at your convenience.

Confidentiality and Use of Data:
Your responses will be kept strictly confidential and will be used solely for improving the DSL. We adhere to the highest ethical standards in our research.

Your expertise and nuanced feedback are vital for the success of this project, and we greatly appreciate your consideration.

Thank you for your time and expertise.
\end{lstlisting}

\subsection{Empirical evaluation}\label{sec:Apdx-empirical-eval}

We take the GPT-4 as our expert in machine empirical evaluation. We built a knowledge augmentation system using Langchain~\footnote{\url{https://www.langchain.com/}} to provide additional domain knowledge to GPT-4. We designed our instruction template following the prompt format of~\citet{zheng2023judging} and utilizing the Plan-and-Solve decomposition idea~\citep{wang2023plan}. The machine evaluation results on DSL of five domains and BioCoder are shown in Table \ref{tab:machine empirical evaluation}.

\begin{table}[ht]
    \centering
    \caption{\textbf{Empirical evaluation by the \ac{qa} expert on the five \acp{dsl} corresponding to five domains and BioCoder}}
    \begin{tabular}{ccc}
        \toprule
         \textbf{Domain}     & \textbf{Opinion}   & \textbf{Confidence} \\ 
        \midrule
        \textbf{BioCoder}     & 4.35  & 3.96   \\ 
        \textbf{InfoBio}     & 4.38  & 3.97     \\ 
        \textbf{Ecology}     & 4.39  & 3.96    \\ 
        \textbf{BioEng}     & 4.41  & 3.97   \\
        \textbf{Medical}     & 4.40  & 3.97  \\ 
        \textbf{Genetics}     & 4.41  & 3.97   \\ 
        \bottomrule
    \end{tabular}
    \label{tab:machine empirical evaluation}
\end{table}

The instruction template is displayed as follows:
\begin{lstlisting}[]
Please act as an impartial judge and evaluate the given developed Domain-Specific Language ([DSL]) according to the [Rating Basis], [Rating Criteria], and related [Domain Knowledges]. Please follow the [Instruction]. Moreover, the [Background] describes something about Domain-Specific Language ([DSL]).
---------------------------------------------------------------
[Background]
Domain experts expect to describe the experimental protocols of the specific domain with the \ac{dsl} programs.
Domain experts expect to be guided by the \ac{dsl} programs step-by-step, where the input, output, and configurations of each step is well-detailed.
Domain experts come without any training on programming.
Domain experts are subjective and their mindsets come from the specific domain.

[Instruction]
The [Rating Basis] presents a series of statements or questions about the [DSL]. Please evaluate your opinion and confidence level in the statements or questions referring to related domain knowledge.
That is, your evaluation should be based primarily [Rating Basis]. You should refer to and use [Domain Knowledge] to help you evaluate. Begin your evaluation by providing a short explanation. After providing your explanation, please rate the [DSL] on [Rating Criteria] by strictly following this format: "Rating: [[<opinion>, <confidence>]]", for example: "Rating: [[3, 4]]".

[Rating Criteria]
For each item, you need give an <Opinion> and <Confidence> rate:
Opinion on a 1-5 Scale (int): Rate each item on a scale where 1 indicates 'Strongly Disagree' and 5 indicates 'Strongly Agree'. The scale is nuanced as follows:
1: Strong Disagreement or Major Issues
2: Disagreement or Notable Concerns
3: Neutral or Mixed Feelings
4: Agreement or Minor Concerns
5: Strong Agreement or Highly Favorable
Confidence in Your Response on a 1-5 Scale (int): Indicate your level of confidence in your response to each item, where 1 is 'Not Confident at All' and 5 is 'Extremely Confident'. The scale implies:
1: Very Low Confidence
2: Low Confidence
3: Moderate Confidence
4: High Confidence
5: Very High Confidence

[Rating Basis]
{}

[DSL]
{}

[Domain Knowledges]
{}

[Output]
Let's first understand/follow the [Instruction] to evaluate the [DSL] and give an explanation about rating.
Then let's give a final rating based on the explanation.
\end{lstlisting}

\texttt{[Rating Basis]} refers to our designed questions or statements.
\begin{lstlisting}[]
"The DSL clearly represents the essential concepts necessary for life sciences experiments."
"All domain concepts included in the DSL are relevant and contribute directly to my tasks in life sciences."
"I can easily express all necessary domain concepts for my experiments using the DSL."
"The DSL does not include complex concepts that are unnecessary for my work in life sciences."
"The DSL seems to have been developed with practical feedback from domain experts, ensuring its relevance to my work."
"The design of the DSL focuses on the tasks I need to accomplish in life sciences, avoiding irrelevant features."
"The DSL is simple and straightforward to understand and use in my life sciences work."
"Learning to use the DSL does not require extensive time or effort, even for those new to this kind of language."
"The tools and features of the DSL are clear and intuitive to use in my everyday tasks."
"The simplicity of the DSL enhances my productivity in conducting life sciences experiments."
"The straightforward nature of the DSL lowers the barrier to its adoption in my professional environment."
"The DSL specifically addresses the unique concepts and needs of my work in life sciences, without unnecessary generalizations."
"All concepts and features in the DSL are directly applicable and useful for my life sciences tasks."
"The DSL avoids overly complex or overly generalized concepts that is uneasy for me to understand, making it more practical for my work."
"The focused nature of the DSL facilitates its quick and successful introduction into my life sciences work."
"The DSL is concise and precise, focusing only on what is necessary for life sciences, without extraneous features."
"The DSL effectively meets the current needs of my work in life sciences without complicating future changes in the domain."
"The number of elements in the DSL is appropriate and manageable for my work in life sciences, making it easy to understand."
"The DSL effectively uses specialized sublanguages (or subsets) for different aspects of life sciences, enhancing clarity and focus."
"Despite having a limited number of elements, the DSL is effective in handling the complex tasks of my life sciences work."
"The DSL's approach to representing different aspects of life sciences work (like structure, behavior, etc.) is clear and well-organized."
"The DSL's limited elements are versatile enough to be applicable across a wide range of life sciences applications."
"Each concept in the DSL is distinct and clearly defined, with no overlap or redundancy in their functions or purposes."
"I find it straightforward to express ideas in my life sciences work using the DSL, as it avoids unnecessary duplication of concepts."
"The DSL allows for consistent modeling of facts and processes in life sciences, without confusion due to redundant concepts."
"Remembering and applying different concepts in the DSL is easy, as it avoids minor variations that could lead to confusion."
"The DSL effectively represents life sciences data and processes without complicating them through redundant concepts."
"The non-redundant design of the DSL positively impacts my workflow in life sciences by providing clarity and efficiency."
"The notations used in the DSL are familiar to me, reflecting those commonly used in my life sciences work."
"Transitioning to using the DSL was easy for me because it adopts notations and terminologies I'm already accustomed to in life sciences."
"I did not have to spend much time learning new syntaxes or notations when starting with the DSL, thanks to its use of familiar concepts."
"The syntax and notation of the DSL are well-suited for the specific tasks I perform in life sciences."
"The DSL effectively integrates languages or notations that are commonly accepted and used in my field (like SQL for database queries, if relevant)."
"The DSL aligns closely with the existing notations and terminologies that are standard in the life sciences domain."
"In the DSL, different elements are clearly distinguishable from one another, enhancing my understanding and ease of use."
"In textual aspects of the DSL, keywords and language elements are positioned in a way that makes the content easy to read and understand."
"The DSL effectively avoids ambiguity in representing different domain concepts, making it simple for me to interpret the models or text."
"The design of the DSL prioritizes readability and understanding for the reader, rather than writing efficiency."
"The DSL allows me to easily add comments to model elements, aiding in documentation and explanation."
"Comments within the DSL models significantly enhance the understandability and clarity of my work."
"The ability to comment on model elements in the DSL simplifies and supports collaborative efforts in my life sciences projects."
"The DSL offers flexibility in commenting styles, such as line comments and block comments for textual languages or annotations for graphical elements."
"Comments in the DSL can be effectively used for detailed documentation, similar to generating HTML pages or Javadoc."
"The DSL encourages or enforces a consistent style of documentation through its comment structure, improving overall model quality."
"The usage conventions defined for the DSL contribute to its clarity and comprehensibility, making it easier for me to use in my life sciences work."
"The DSL strikes a good balance between providing guidance through usage conventions and allowing flexibility in how I use it."
"The established conventions in the DSL aid in maintaining the quality and consistency of my work."
"Following the usage conventions in the DSL is straightforward and does not add unnecessary complexity to my tasks."
"The conventions in the DSL, such as the notation of identifiers and the order of elements, enhance the overall comprehensibility of the language."
"The usage conventions of the DSL positively impact its overall usability and practicality in my life sciences work."
\end{lstlisting}

\paragraph{Rationale of scoring}

We showcase some non-cherry-picked scoring rationales given by both human expert and our \ac{qa} expert in.

\begin{table*}[h!]
    \centering
    \caption{\textbf{Showcases of scoring rationales.}(the corresponding scoring is attached \textbf{[Opinion, Confidence]})}
    \small 
    \begin{tabularx}{\linewidth}{X|X|X}
    \toprule
    Criteria     &  human expert & \ac{qa} expert \\
    \midrule
    "The DSL clearly represents the essential concepts necessary for life sciences experiments." &  The DSL precisely covers key life sciences experiments, enhancing protocol clarity and accuracy. \textbf{[5, 4]} & The DSL for life sciences effectively encapsulates key experimental actions—ASPIRATE, ADD, CLEAN, COMBINE, PUT—covering essential concepts from aspirating liquids to placing items. Its design and domain-specific actions provide a comprehensive framework for describing experiments, reflecting its adequacy for the life sciences domain. \textbf{[5, 4]}\\
    \midrule
    "The DSL allows for consistent modeling of facts and processes in life sciences, without confusion due to redundant concepts." & The DSL clarifies life sciences protocols, potentially outperforming traditional methods with its precision. \textbf{[4, 4]} & The DSL effectively captures key experimental actions in life sciences, offering clarity with specific patterns and examples for each action. Rated with high confidence, it ensures precise protocol definition, minimizing confusion and redundancy. Further analysis could enhance its validation against existing standards. \textbf{[4, 4]}\\
    \midrule
    "The DSL is simple and straightforward to understand and use in my life sciences work." & The DSL simplifies life sciences protocols with clear commands and examples, though it may benefit from expanded flexibility. \textbf{[4, 4]} & The DSL is effectively designed for life sciences, offering domain-specific commands like ASPIRATE and ADD with clear examples, facilitating experimental protocols without programming knowledge. While comprehensive and user-friendly, potential areas for additional commands or flexibility could further enhance its utility. \textbf{[4, 4]}\\
    \midrule
    "The DSL's limited elements are versatile enough to be applicable across a wide range of life sciences applications." & The DSL streamlines essential life sciences experiments but may fall short in specialized contexts, justifying a cautious yet positive rating of 4. \textbf{[4, 4]} & The DSL effectively guides life sciences experts through standard experiments with commands like ASPIRATE and ADD, balancing simplicity and specificity. Rated at 4, it offers a solid foundation for common protocols but may need expansion for more specialized experiments. \textbf{[4, 4]}\\
    \bottomrule
    \end{tabularx}
    \label{tab:my_label}
\end{table*}

\subsection{Ontology extraction}

We take a zero-short \ac{ie} tool as our backbone to extract ontologies from protocols. We add the domain-related definition to the Stage-I prompt of RE task in ChatIE, which is defined as follows: \texttt{...In this task, an entity type may be a 'OpCode' (operations, a one-word verb, like ADD and REMOVE); 'REG' (reagents taking part in an operation, like cells and MDDC culture media); 'COND' (conditions of executing an operation, like <temperature> 37C, <time> 30min, <device> a small scissor and <container> PCR tubes)...}.

In addition, we present the entity relation mapping dictionary as follows: 
\begin{lstlisting}[]
{
 'is concurrent with': ['OpCode', 'OpCode'],
 'is instruction of': ['OpCode', 'REG/COND'],
 'is predecessor of': ['OpCode', 'OpCode'],
 'is product of': ['REG', 'OpCode'],
 'is reaction acceleration of': ['Acceleration', 'OpCode'],
 'is reaction centrifugal force of': ['Centrifugal Force', 'OpCode'],
 'is reaction condition of': ['COND', 'OpCode'],
 'is reaction container of': ['Container', 'OpCode'],
 'is reaction density of': ['Density', 'OpCode'],
 'is reaction device of': ['Device', 'OpCode'],
 'is reaction energy of': ['Energy', 'OpCode'],
 'is reaction flow rate of': ['Flow Rate', 'OpCode'],
 'is reaction force of': ['Force', 'OpCode'],
 'is reaction frequency of': ['Frequency', 'OpCode'],
 'is reaction iteration count of': ['Iteration Count', 'OpCode'],
 'is reaction pressure of': ['Pressure', 'OpCode'],
 'is reaction rotation of': ['Rotation', 'OpCode'],
 'is reaction speed of': ['Speed', 'OpCode'],
 'is reaction temperature of': ['Temperature', 'OpCode'],
 'is reaction time of': ['Time', 'OpCode'],
 'is reaction voltage of': ['Voltage', 'OpCode'],
 'is reagent acidity of': ['Acidity', 'OpCode'],
 'is reagent coating of': ['Coating', 'OpCode'],
 'is reagent concentration of': ['Concentration', 'OpCode'],
 'is reagent density of': ['Density', 'OpCode'],
 'is reagent length of': ['Length', 'OpCode'],
 'is reagent mass of': ['Mass', 'OpCode'],
 'is reagent medium of': ['Medium', 'OpCode'],
 'is reagent of': ['REG', 'OpCode'],
 'is reagent quantity of': ['Quantity', 'OpCode'],
 'is reagent size of': ['Size', 'OpCode'],
 'is reagent thickness of': ['Thickness', 'OpCode'],
 'is reagent volume of': ['Volume', 'OpCode'],
 'is successor of': ['OpCode', 'OpCode']
}
\end{lstlisting}

\section{Constraint utility assessment details}

\subsection{Protocol source}

In order to better analyze the subsequent \acp{dsl}, we created a protocol dataset of five domains in the life sciences in this paper, where all the protocols are not pre-existing protocols, but NOVEL ones, designed by ourselves, and of course most of these designed protocols are unpublished and written with reference to a large number of existing protocols.

\subsection{Protocol constraint}

The protocol dataset we built covers both syntactic constraints and semantic constraints. We categorized these two main features into multiple types and built specific and typical protocols for each subclass, which can encompass most of the experimental scenarios.

\paragraph{Syntactic constraint}

Syntactic constraints can be primarily divided into four categories: the first category, the imperative model, includes loop, branch, jump, memory (allocation and de-allocation), function/(sub)procedure, and logical test. The second category, the type system, encompasses integer types, real/floating types, set types, scientific types, reagent types, and container types. The third category, concurrent, contains data parallel. The fourth category is the reactive model, which responds to a certain event.

\begin{lstlisting}[]
{
Loop constraint
 Cell culture: 
 1. Prepare medium: prepare appropriate medium according to cell type. 
 2. Inoculate: Add cell suspension to culture flasks containing culture medium.
 3. Continuous monitoring: Observe the cell growth status and density periodically. Repeat the culture as long as the cells have not reached the harvested density. 
 4. Harvest cells: When the cells reach the expected density, harvest the cells for the next experiment or for passaging culture.
}
{
Branch constraint
Brown Adipocyte Respiration Protocol: Prepare cells as described in the first two steps of the intact brown adipocyte respiration protocol. On the day of the experiment, thaw 1x MAS and pre-made substrate solutions. Prepare 10x solutions of inhibitors in 1x MAS. If using fatty acid substrates, prepare them freshly as described. Soak the cartridge for at least 3 hours before loading. Load specified volumes into ports A-D and calibrate the cartridge. Prepare 7.5 nM PMP in 1x MAS, gently wash cells with 1x MAS once, completely evacuate the wash, and add 450 µl 1x MAS with 7.5 nM PMP. Perform the Seahorse assay with 2-3 measurements per condition, ensuring not to exceed 1 hour before injecting Antimycin A to maintain mitochondrial integrity.
}
{
Memory constraint
Bile Processing Protocol: Collect bile in sterile bottles from patients with biliary tubes. Centrifuge at 3000g for 10 minutes at 4°C to remove sediment and debris. Aliquot the clear bile into 1 ml volumes in microcentrifuge tubes. Store the aliquots at -20°C for future use.
}
{
Function procedure constraint
Isolation of G4 Deletion Alleles in C. elegans: Clone out dog-1 animals onto plates, optionally freeze half as backup. Rinse worms in M9, transfer to PCR plates, add lysis buffer, and proceed with lysis and proteinase K inactivation. Perform nested PCR on lysis mix to identify positive populations for germline mutations. Analyze PCR products on agarose gel. For positive populations, chunk corresponding plate, grow to starvation, and repeat the process to confirm deletions. Perform nested PCR on progeny to confirm homozygosity of deletion alleles. Sequence PCR product for deletion characterization. Backcross new strain to N2 to clean genetic background.
}
{
Parallel constraint
Centrifugation and RNA Isolation from Serum or Plasma: Start with 1 ml of serum or plasma for each replicate. First centrifugation: Bring volume up with PBS and centrifuge all samples at once for 90 min at 100,000 x g. Second centrifugation: After discarding supernatant and resuspending pellets, centrifuge again for 70 min at 100,000 x g. Parallel processing: Process multiple samples through two rounds of centrifugation and subsequent RNA isolation, highlighting batch processing of samples.
}
{
Reactive model constraint
Cell Growth and Transfection Response: Thaw and culture cells until 80% confluency. Response to 80% confluency: Split cells for further growth or prepare for freezing. Transfect cells at low density, then change media based on purpose (growth or differentiation). Response to transfection: Monitor for complete differentiation, adjusting care based on confluency or media type."
}
\end{lstlisting}

\paragraph{Semantic constraint}

Semantic constraints mainly include two types: (i) Use of undefined action; (ii) Incomplete parameter, where the first category is required parameters do not exist, and the second category is Required parameter under-specified (is not grounded to the granularity for execution).

\subsection{Prompt engineering for utility assessment}

Following methods suggested by recent research \citep{gao2023pal,zhang2023causal}, we prompt the LLM with examples of DSL syntax and semantics, directing it to translate procedural texts into corresponding programs. To ensure a fair comparison and to underscore the plug-and-play capability of the \acp{dsl} designed by \ad, only minimal prompt engineering is applied in protocol processing. The resulting prompt text under the ideas is demonstrated as follows.

\begin{lstlisting}[]
You are an expert in life science and computer science. Now you are prompted with a grammar of programming language defined by production rules, several experiment steps described in natural language, and a construct (which is the left part in the production rule). Your task is to determine whether the natural language description consists of parts that can be parsed using this production rule. If the natural language string can be parsed, please output "Yes", otherwise, output "No".

The production rule:
for (initialization; condition; increment) statement

The nature language experimental protocol:
Study design: Timing: 3 days for protocol development. IRB/ethics review and revision can take 4-8 weeks. Repeat with number from 1 to 11:1. Define the crisis situation.2. Select existing algorithms for study.3. Define hypothetical triage algorithms for study.4. Select primary and secondary clinical endpoints.5. Determine the clinical data needed.6. Select the patient cohort for study.7. Assess availability of necessary clinical data.8. Adapt the triage algorithm scoring scheme.9. Submit the proposed study to the IRB or ethics panel. Calculating priority scores: Timing: 3 weeks to 3 months. Repeat with number from 1 to 8:1. Create a case report form.2. Select a data management system.3. Select the method of data entry. 4. Perform pilot data acquisition. 5. Complete the case report form for all patients.6. Apply the triage algorithms and calculate priority scores. Testing algorithm accuracy Timing: 1 day.1. Determine the accuracy of the priority scores Simulation of clinical decision-making: Selection from a smaller group Timing: 1 day. Repeat with number from 1 to 4:1. Format input files.2. Run CSC script.3. Assess bootstrap analysis output. Sensitivity analyses: Timing: 1 week. Repeat with number from 1 to 4:1. Test the effect of data processing methods.2. Test the effect of patient characteristics. 3. Test the effect of triage algorithm components.
yes

The production rule:
for (initialization; condition; increment) statement

The nature language experimental protocol:
Mitochondria Sonication Experiment:Repeat sonication of mitochondria in a Branson 450 sonicator using aconitase buffer (50 mM Tris, 30 mM sodium citrate, 0.5 mM MnCl2, 0.2 mM NADP, pH 7.3) four times, each for 15 seconds.Monitor the citrate to alpha-ketoglutarate conversion at 340 nm at 25°C, utilizing 2 units/ml of isocitrate dehydrogenase in 50 mM Tris, 1 mM cysteine, 1 mM sodium citrate, 0.5 mM MnCl2 at pH 7.4. Reactivate aconitase with 2 mM dithiothreitol and 0.2 mM ferrous ammonium sulfate for 5 minutes, then repeat the enzymatic activity assay once.

Task:
Please use a JSON format to describe the protocol (Only output a json).

Note: 
1. Read the protocol carefully.
2. Choose items and conditions from the protocol to describe the protocol.
3. Your desired output format resembles this: {"opcode": [['Datatype', 'Data'], ... , ["output", "Data"]]}. In this format, "Data" = None indicates a missing value.
4. You must specify the output in json, and only one output in json is allowed. 

Example:
GROW:[['REG', 'GFP-fusion'], ["REG", "solid media"], ["Device", "spinning disk confocal microscope"]] -> Z-stack images
Output: {"GROW": [['REG', 'GFP-fusion'], ["REG", "solid media"], ["Device", "spinning disk confocal microscope"], ["output", "Z-stack images"]]}
\end{lstlisting}

\end{document}